\UseRawInputEncoding
\documentclass[letterpaper, 10 pt, conference]{ieeeconf}  

\IEEEoverridecommandlockouts                              

\usepackage{amsmath, bm} 
\usepackage{amssymb}
\usepackage{pdfpages}
\usepackage{subcaption}
\usepackage{siunitx}
\usepackage{xcolor}
\usepackage{url}
\usepackage[ruled,linesnumbered,hangingcomment]{algorithm2e}
\usepackage{algpseudocode}
\usepackage{physics}
\usepackage{mathtools}
\usepackage[]{hyperref}
\usepackage[nolist]{acronym}
\usepackage{cite}
\usepackage[english]{babel}
\usepackage[utf8]{inputenc}

\def\BibTeX{{\rm B\kern-.05em{\sc i\kern-.025em b}\kern-.08em
    T\kern-.1667em\lower.7ex\hbox{E}\kern-.125emX}}

\SetKwInOut{Parameters}{Parameters}

\title{\LARGE \bf
Cooperative Indoor Exploration Leveraging a Mixed-Size UAV Team with Heterogeneous Sensors}

\usepackage{scalerel}
\usepackage{tikz}
\usetikzlibrary{svg.path}
\definecolor{orcidlogocol}{HTML}{A6CE39}
\tikzset{
  orcidlogo/.pic={
    \fill[orcidlogocol] svg{M256,128c0,70.7-57.3,128-128,128C57.3,256,0,198.7,0,128C0,57.3,57.3,0,128,0C198.7,0,256,57.3,256,128z};
    \fill[white] svg{M86.3,186.2H70.9V79.1h15.4v48.4V186.2z}
    svg{M108.9,79.1h41.6c39.6,0,57,28.3,57,53.6c0,27.5-21.5,53.6-56.8,53.6h-41.8V79.1z M124.3,172.4h24.5c34.9,0,42.9-26.5,42.9-39.7c0-21.5-13.7-39.7-43.7-39.7h-23.7V172.4z}
    svg{M88.7,56.8c0,5.5-4.5,10.1-10.1,10.1c-5.6,0-10.1-4.6-10.1-10.1c0-5.6,4.5-10.1,10.1-10.1C84.2,46.7,88.7,51.3,88.7,56.8z};
  }
}

\newcommand\orcidicon[1]{\href{https://orcid.org/#1}{\mbox{\scalerel*{
        \begin{tikzpicture}[yscale=-1,transform shape]
          \pic{orcidlogo};
        \end{tikzpicture}
}{|}}}}

\author{
    Michaela Cihl\'{a}\v{r}ov\'{a}$^*$,
    V\'{a}clav Pritzl$^{*{\orcidicon{0000-0002-7248-6666}}}$,
    Martin Saska$^{*{\orcidicon{0000-0001-7106-3816}}}$%
    \thanks{This work was funded by CTU grant no. SGS23/177/OHK3/3T/13, by the European Union under the project Robotics and Advanced Industrial Production (reg. no. CZ.02.01.01/00/22\_008/0004590) and by the Czech Science Foundation (GAČR) under research project no. 23-07517S.
    $^*$Authors are with the Department of Cybernetics, Faculty of Electrical Engineering, Czech Technical University in Prague, 166 36 Prague 6, {\tt\footnotesize\{\href{mailto:cihlami1@fel.cvut.cz}{cihlami1}|\href{mailto:pritzvac@fel.cvut.cz}{pritzvac}|\href{mailto:martin.saska@fel.cvut.cz}{martin.saska}\}@fel.cvut.cz}}%
}

\begin{acronym}
\acro{A-LOAM}[A-LOAM]{Advanced implementation of LOAM}
\acro{CTU}[CTU]{Czech Technical University}
\acro{DARPA}[DARPA]{Defense Advanced Research Projects Agency}
\acro{DOF}[DOF]{degree-of-freedom}
\acro{FFD}[FFD]{Fast Frontier Detector}
\acro{FOV}[FOV]{Field of View}
\acro{GNSS}[GNSS]{Global Navigation Satellite System}
\acro{GPS}[GPS]{Global Positioning System}
\acro{IMU}[IMU]{Inertial Measurement Unit}
\acro{IR}[IR]{infrared}
\acro{LiDAR}[LiDAR]{Light Detection and Ranging}
\acro{LKF}[LKF]{Linear Kalman Filter}
\acro{LOAM}[LOAM]{LiDAR Odometry and Mapping}
\acro{MBZIRC}[MBZIRC]{Mohamed Bin Zayed International Robotics Challenge}
\acro{MCF}[MCF]{Minimum-cost flow}
\acro{MPC}[MPC]{Model predictive control}
\acro{MRS}[MRS group]{Multi-robot Systems group}
\acro{MRS system}[MRS UAV system]{Multi-robot Systems Group UAV system}
\acro{MRTA}[MRTA]{Multi-Robot Task Allocation}
\acro{NAV}[NAV]{Nano Unmanned Aerial Vehicles}
\acro{NBV}[NBV]{Next-Best-View}
\acro{NTP}[NTP]{Network Time Protocol}
\acro{POI}[POI]{Point of Interest}
\acroplural{POI}[POIs]{Points of Interest}
\acro{pUAV}[pUAV]{primary Unmanned Aerial Vehicle}
\acro{RGB}[RGB]{Red-Green-Blue}
\acro{RGBD}[RGBD]{Red-Green-Blue-Depth}
\acro{ROS}[ROS]{Robot Operating System}
\acro{RRT}[RRT]{Rapidly Exploring Random Tree}
\acro{SLAM}[SLAM]{Simultaneous Localization And Mapping}
\acro{sUAV}[sUAV]{secondary Unmanned Aerial Vehicle}
\acro{SubT}[SubT]{Subterranean}
\acro{TSP}[TSP]{Traveling Salesman Problem}
\acro{UAV}[UAV]{Unmanned Aerial Vehicle}
\acro{UGV}[UGV]{Unmanned Ground Vehicle}
\acro{VIO}[VIO]{Visual Inertial Odometry}
\acro{WFD}[WFD]{Wavefront Frontier Detector}
\end{acronym}

\newcommand{\PREPRINTYEAR}{2024}
\newcommand{\PUBLISHEDIN}{IEEE}
\newcommand{\PUBLISHEDINSHORT}{IEEE CASE 2024}
\newcommand{\DOI}{10.1109/CASE59546.2024.10711365} 

\usepackage[placement=top,vshift=-7]{background}
\SetBgScale{1.0}
\SetBgContents{\PUBLISHEDINSHORT. ACCEPTED VERSION - DO NOT DISTRIBUTE. \href{https://doi.org/\DOI}{DOI \DOI}}
\SetBgColor{black}
\SetBgAngle{0}
\SetBgOpacity{1.0}

\begin{document}

\thispagestyle{empty}
\onecolumn
{
  \topskip0pt
  \vspace*{\fill}
  \centering
  \LARGE{%
    \copyright{} \PREPRINTYEAR~\PUBLISHEDIN\\\vspace{1cm}
    Personal use of this material is permitted.
    Permission from IEEE~must be obtained for all other uses, in any current or future media, including reprinting or republishing this material for advertising or promotional purposes, creating new collective works, for resale or redistribution to servers or lists, or reuse of any copyrighted component of this work in other works.}
    \vspace*{\fill}
}
\NoBgThispage
\twocolumn          	
\BgThispage

\maketitle
\thispagestyle{empty}
\pagestyle{empty}

\setlength{\textfloatsep}{5pt}

\begin{abstract}
Heterogeneous teams of \acp{UAV} can enhance the exploration capabilities of aerial robots by exploiting different strengths and abilities of varying \acp{UAV}.
This paper presents a novel method for exploring unknown indoor spaces with a team of \acp{UAV} of different sizes and sensory equipment. We propose a frontier-based exploration with two task allocation strategies: a greedy strategy that assigns \acp{POI} based on Euclidean distance and \ac{UAV} priority and an optimization strategy that solves a minimum-cost flow problem. The proposed method utilizes the SphereMap algorithm to assess the accessibility of the \acp{POI} and generate paths that account for obstacle distances, including collision avoidance maneuvers among \acp{UAV}. The proposed approach was validated through simulation testing and real-world experiments that evaluated the method's performance on board the \acp{UAV}.

The paper is supported by the multimedia materials available at
\url{https://mrs.felk.cvut.cz/case2024exploration}.

\end{abstract}

\acresetall

\vspace{-0.5em}
\section{\textsc{Introduction}}

Heterogeneous teams of UAVs with different roles, each equipped with different sensors, can handle diverse challenges by leveraging different capabilities. By combining \acp{UAV} with different sensory modes, endurance levels, and maneuvering skills, these teams can better adapt to complex and dynamic environments.

One such application is mapping indoor areas with narrow passages and openings. A team consisting of a bigger \ac{UAV} with precise sensors and high computing capabilities and one or multiple smaller dependent \acp{UAV} capable of fitting through narrow entrances at the cost of sensor accuracy is especially suitable for such complex environments. The smaller \acp{UAV} can explore locations that are hard to reach, while the bigger \ac{UAV} uses its onboard sensors to create global maps, synchronize the fleet, and guide the exploration. In addition to more reliable and precise localization of the entire team in the global map, this approach protects the more expensive \ac{UAV} while the cheaper ones perform riskier tasks. 

The incorporation of heterogeneity within \ac{UAV} teams offers numerous benefits for exploratory missions. Yet, it also introduces several challenges that need to be addressed: how to allocate tasks among the \acp{UAV} according to their individual capabilities, how to coordinate the \acp{UAV} to achieve a common goal while respecting their constraints, and how to balance the trade-off between finding better solutions and ensuring real-time computation on board the \acp{UAV}. These challenges require careful attention to ensure that the benefits of heterogeneity outweigh the costs and do not compromise the overall effectiveness of the exploration mission.

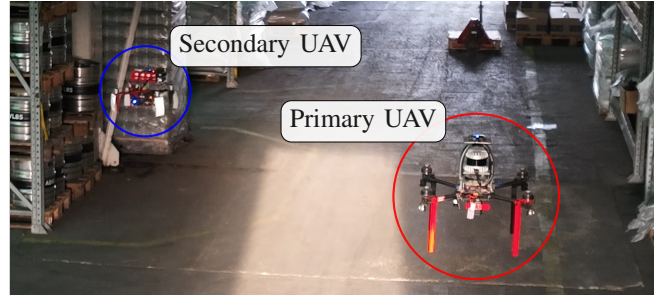
\begin{figure}[t]
\centering
\begin{tikzpicture}
    \node[anchor=north west, inner sep=0, draw=black] (a) at (0, 0) {
    \includegraphics[width=.99\linewidth, trim={0mm 4cm 0mm 0mm},clip]{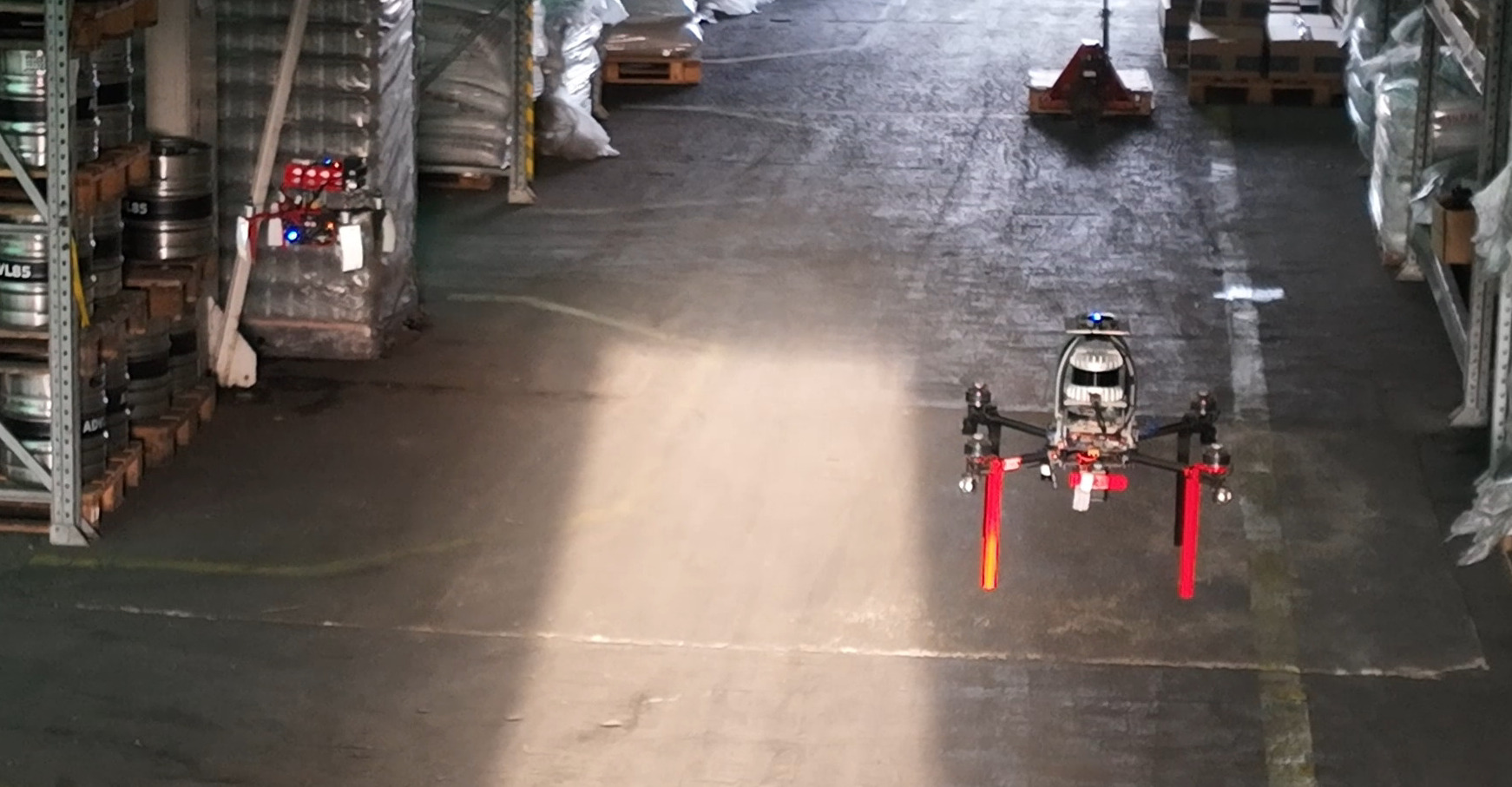}
    };
    \draw[opacity=0.9, color=blue, thick] (1.8, -1.2) circle [radius=0.6];
    \draw[opacity=0.9, color=red, thick] (6.2, -2.6) circle [radius=1.1];
     \node[fill=white, rounded corners, draw=black, opacity=0.9] (label1) at (4.7, -1.6) {
      \color{black}{Primary UAV}
    };
    \node[fill=white, rounded corners, draw=black, opacity=0.9] (label2) at (3.4, -0.6) {
      \color{black}{Secondary UAV}
    };
\end{tikzpicture}
\caption{\acs{UAV} platforms used for experimental verification of the proposed method in real-world conditions: larger primary \acs{UAV}, and smaller secondary \acs{UAV}.}
\label{fig:uav-paltform}
\end{figure}

In this paper, we propose an exploration strategy using a team of two closely cooperating \acp{UAV}; a larger, more-capable \ac{pUAV}, and a smaller, cheaper \ac{sUAV} (see Fig. \ref{fig:uav-paltform}). The intentional variation of \ac{UAV} capabilities within the team enables it to quickly and accurately map spacious areas by the \ac{pUAV} and simultaneously map confined spaces using the \ac{sUAV}. This approach underscores the utility of heterogeneity in enhancing the efficiency and effectiveness of \ac{UAV}-based exploration.

\vspace{-0.5em}
\section{\textsc{Problem Statement}}
\label{section:problem-statement}

We address the challenge of autonomous exploration of complex, \ac{GNSS}-denied environments through cooperative deployment of multiple \acp{UAV} featuring different physical characteristics and sensory capabilities.
In this context, two \acp{UAV} are considered. The first is a larger \ac{pUAV}, equipped with high computation power and a 3D \ac{LiDAR} sensor with substantial range and high-precision data acquisition capabilities. The second is a comparatively smaller \ac{sUAV}, intentionally designed to navigate through narrow spaces. However, it is limited to a \ac{RGBD} camera with considerably reduced \ac{FOV} and lower-quality mapping functionality. The \acp{UAV} are capable of mutual communication over a wireless network, and the algorithms are designed to run fully on board the \acp{UAV} with no external computational resources utilized.

The task is to optimize the collaborative efforts of these two \acp{UAV} to explore and map a priori-unknown environment where physical constraints and differing sensory inputs impact their exploration capabilities. This can be divided into the following components:
\begin{enumerate}
    \item \textbf{Selecting \acfp{POI}} - the challenge of selecting relevant points within the environment to guide the \acp{UAV}' exploration efforts.

    \item \textbf{Accessibility Problem} - the consideration of accessibility constraints, particularly indoors with confined spaces, ensuring that the selected points are reachable by a specific \ac{UAV} (\ac{pUAV} or \ac{sUAV}).

    \item \textbf{Task Allocation} - the problem of distributing specific exploration tasks among the \acp{UAV} efficiently to maximize exploration coverage.

    \item \textbf{Planning} - path planning strategies for the \acp{UAV} to navigate from their current location to the selected points of interest.

    \item \textbf{Obstacle Avoidance} - the \acp{UAV} must safely navigate around static obstacles like walls and furniture, as well as dynamically moving objects (other \acp{UAV}).
\end{enumerate}

In the method description, vectors are denoted with bold lowercase letters, matrices with bold uppercase italic letters, and frames of reference with uppercase upright letters. Sets and sequences are denoted by uppercase calligraphic letters. The transformation matrix describing the transition from frame A to frame B is represented as $\prescript{\mathrm{B}}{\mathrm{A}}{\boldsymbol{T}} \in SE(3)$. Let $\prescript{\mathrm{A}}{}{\boldsymbol{x}} \in \mathbb{R}^3$ be a 3D position vector in frame A, and let $\prescript{\mathrm{A}}{}{\mathcal{P}_B}$ be a sequence of \ac{UAV} reference poses $(\prescript{\mathrm{A}}{}{\boldsymbol{x}_i}, \prescript{\mathrm{A}}{}{\phi_i})$, with position $\prescript{\mathrm{A}}{}{\boldsymbol{x}_i} \in \mathbb{R}^3$ and heading/yaw orientation $\prescript{\mathrm{A}}{}{\phi_i} \in [-\pi, \pi]$, for \ac{UAV} $B$ in reference frame A.


\section{\textsc{Related Work}}
\label{section:related}

The exploration of unknown spaces has been addressed mainly through sampling-based and frontier-based approaches or their combinations.
Sampling-based approaches, linked with the \ac{NBV} method~\cite{related:next-best-view-orig, related:sa-rrt-orig, related:sa-rrt-based}, involve randomly sampling candidate viewpoints and selecting the one with the largest information gain.
Frontier-based approaches~\cite{related:frontier-orig, related:me-front-old, related:front-fuel} rely on identifying the frontiers, i.e., the boundaries between the known and unknown areas.
Robots move toward these frontiers to gather information about the unexplored regions.
In this work, a frontier-based approach is utilized as the accessibility of a specific frontier can be separately evaluated for each \ac{UAV} and utilized for efficient goal assignment among the heterogeneous \ac{UAV} team.

Our work focuses on creating a real-time system that is able to navigate safely through narrow passages, which requires fast and safety-aware planning. We obtain crucial information about the environment by using a volumetric occupancy grid such as the OctoMap \cite{octomap}. However, long-distance planning on such a structure is time-consuming. In \cite{related:he-multi-sensor}, the authors address this issue by creating a cache of the collision checking results. This reduced the planning time, but it would be inefficient for mixed-sized teams with different constraints. Another approach is to use a simplified, topological representation of the environment \cite{gbplanner, spartan, spheremap}. 
Some of these methods, however, do not focus on the safety of the \ac{UAV} \cite{gbplanner} or are not suitable for large-scale exploration \cite{spartan}.
SphereMap approach \cite{spheremap} fills free space using intersecting spheres with a predefined minimal radius and creates a graph connecting the centers of the intersecting spheres. It is built continually on board a UAV during the mission, and stores precomputed paths for even faster planning. 
In our approach, SphereMap provides fast planning and computationally efficient checking of viewpoint reachability by differently-sized \acp{UAV}.

The research of multi-\ac{UAV} teams and heterogeneous systems for exploration was supported by \ac{DARPA} \ac{SubT} Challenge\cite{related:me-subt-mrs, related:he-graph-global-map, related:darpa-fill-2, related:darpa-fill-1, related:he-multi-sensor}. An approach using a topological representation of free space is proposed in \cite{related:he-graph-global-map}. It incorporates a team of one \ac{UAV} and one \ac{UGV}, but they explore the area separately, only occasionally sharing locally created maps.
In \cite{related:he-multi-sensor}, the authors focus on the effective use of both range and vision sensing modalities in multi-\ac{UAV} teams. Nevertheless, the experiments were executed with a team consisting of multiple identical \acp{UAV} with the same set of sensors and also flying independently in their workspace.

In \cite{hetero-inspection}, the authors present a methodology for building inspection employing a team of \acp{UAV}. The approach utilizes \ac{LiDAR}-equipped drones for initial exploration phases and camera-equipped drones for detailed inspection tasks. However, the research does not address variations in \ac{UAV} sizes. Moreover, the \acp{UAV} equipped with cameras are restricted to navigating environments previously mapped by \ac{LiDAR} drones. 

Similarly, the advantages of a heterogeneous \ac{NAV} team with various equipment are studied in \cite{related:he-navs-mapping}. \ac{LiDAR}-\acp{NAV} are tasked to perceive the surrounding environment, whereas an Edge-\ac{NAV} collects data from \ac{LiDAR}-\acp{NAV} and constructs the map.
The study’s focus is on creating a map representation based on OctoMap that is optimized for scenarios with limited memory capacity and offloading computational tasks to another \ac{NAV}, but the experiments were again conducted with same-sized \acp{UAV}. 

To achieve effective exploration, the operation area can be divided using Voronoi graph \cite{related:assign-voronoi-com} or grid-based space decomposition \cite{related:me-decent-racer}. However, such approaches are not desirable in the case of a mixed-size \ac{UAV} team exploring a complex indoor space. Each area may contain spaces reachable by only the smaller \ac{UAV}, along with large open areas that are quickly mappable by the larger \ac{UAV} with better sensors. Therefore, the exploration method needs to distinguish between such spaces and assign them to the appropriate \ac{UAV} with corresponding capabilities.

\begin{figure*}
    \centering
    \includegraphics[width=\linewidth]{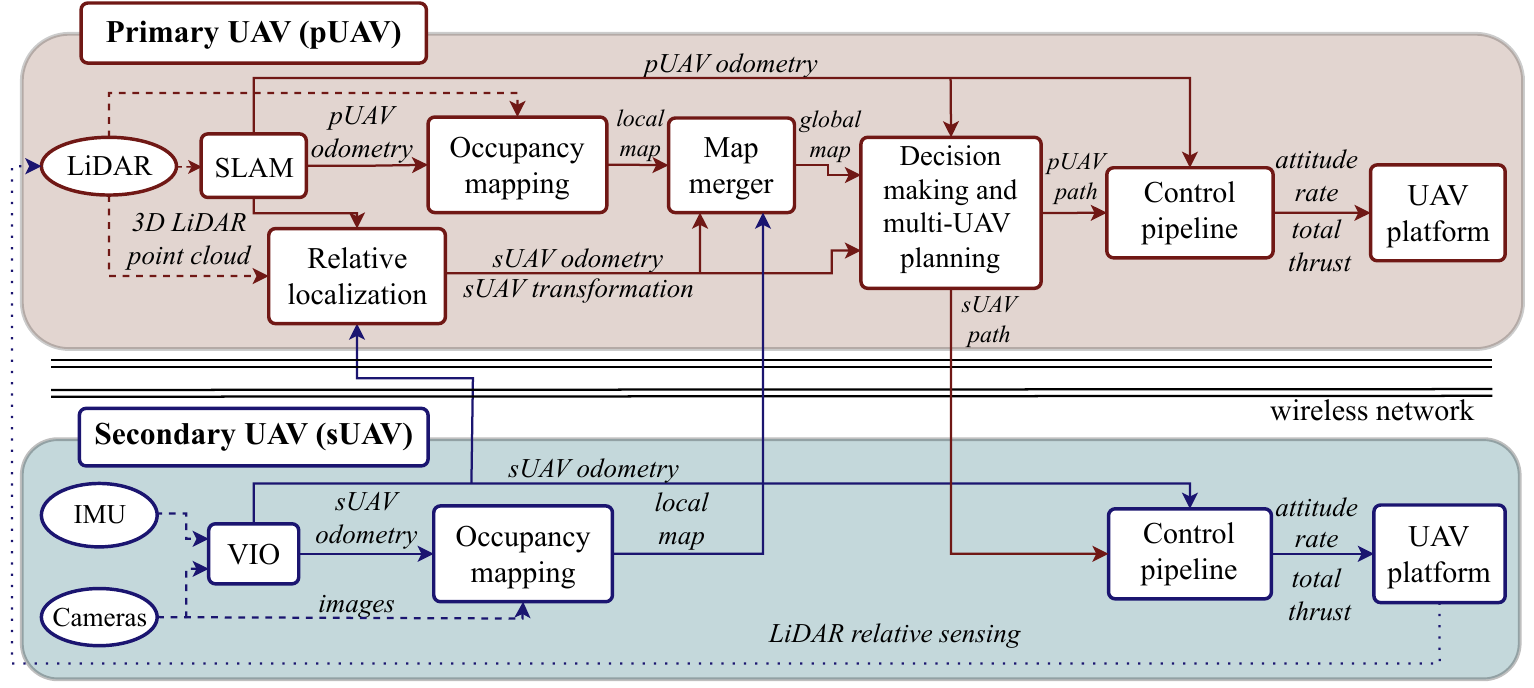}
    \caption{Diagram of the multi-\acs{UAV} software pipeline.}
    \label{fig:diagram-system}
    \vspace{-1.0em}
\end{figure*}

The contribution of this paper is summarized as a novel algorithm for frontier-based cooperative exploration of complex indoor spaces leveraging \acp{UAV} of different sizes and sensory equipment. The proposed approach was evaluated in realistic simulations and on board real \ac{UAV} hardware with no external computational resources required, which proved its viability and real-time performance. To the best of our knowledge, no such cooperative exploration method leveraging heterogeneity among mixed-sized \acp{UAV} has been proposed yet.

\vspace{-0.5em}
\section{\textsc{Multi-UAV Exploration Framework}}

The proposed approach is built on a software pipeline visualized in Fig. \ref{fig:diagram-system}. The \ac{pUAV} utilizes a 3D \ac{LiDAR} \ac{SLAM} algorithm for odometry computation and self-localization. The \textit{Occupancy mapping} generates, using the OctoMap \cite{octomap} approach, a local occupancy map $\prescript{\mathrm{P}}{}{\mathcal{M}_P}$ in \ac{pUAV}'s body frame P. \textit{Relative localization} of the \ac{sUAV} is achieved by identifying the \ac{sUAV} within the \ac{LiDAR} data and fusing the detections with \ac{sUAV} \ac{VIO} odometry in \ac{sUAV}'s body frame S received over a wireless network. A detailed description of the relative localization method is in \cite{rl1}. The \textit{Map merger} algorithm creates a global occupancy map $\prescript{\mathrm{G}}{}{\mathcal{M}}$ in a common global frame G, facilitating high-level planning and decision-making. 

The \ac{sUAV} constructs a local occupancy map from the depth camera data and utilizes \ac{VIO} for its self-localization. The \ac{sUAV} \ac{VIO} odometry and local occupancy map $\prescript{\mathrm{S}}{}{\mathcal{M}_S}$ are sent to the \ac{pUAV} for collaborative control and planning.

The \textit{Control pipeline} on board each UAV follows the desired UAV paths in the corresponding local reference using feedback control utilizing the corresponding self-localization data. Detailed information about the control pipeline is described in \cite{bib:mrs-system}.

The novel \textit{Decision-making and multi-\ac{UAV} planning} algorithms are defined in the global frame G. Without loss of generality, we set the global frame G to be equivalent to the \ac{pUAV} \ac{SLAM} frame P for the specific case of a single \ac{LiDAR}-equipped \ac{pUAV} cooperating with a single camera-equipped \ac{sUAV}. For better clarity, the superscript G, denoting the global frame, is omitted in the description of the algorithms unless its usage is necessary to prevent ambiguity.

The proposed exploration algorithm runs on board the pUAV without any external computational resources and coordinates the behavior of the entire UAV team. The algorithm consists of the following submodules, each running in parallel in a separate thread:

\begin{enumerate}
    \item \textbf{Decision making} - A frontier detection algorithm identifies unexplored regions. The system assigns goals to individual \acp{UAV}, based on the goal's accessibility and distance, optimizing the distribution of exploration efforts across the fleet.
    \item \textbf{Path planning} - The second thread handles the planning of paths to already selected \acp{POI}. It identifies \acp{UAV} completing their current task and requiring a new path. This ensures effective and continuous exploration of the unknown space.
    \item \textbf{Collision avoidance} - The third thread addresses the crucial aspect of collision avoidance among \acp{UAV}. This algorithm continuously monitors the spatial dynamics of the two \acp{UAV} and employs safety measures if needed.
\end{enumerate}

\subsection{Selecting \acfp{POI} and Accessibility}
\label{section:impl_front}

 Frontiers are identified as unoccupied leaf nodes within the exploration area and next to unknown space. The algorithm clusters nearby frontiers based on Euclidean distance and selects the \acp{POI} from these clusters based on their centroids. In complex environments, additional random samples are drawn from each cluster for exploration. The frontier-based algorithm generates the set of \acp{POI} $\mathcal{G}$ used by the Multi-Robot Task Allocation.

The key component of our work is ensuring the safety of drone operations. As mentioned in Sec. \ref{section:related}, we propose an approach that utilizes the SphereMap.

\begin{figure}
\centering
\begin{tikzpicture}
    \node[anchor=north west,inner sep=0] (a) at (0, 0)
    {
      \includegraphics[width=0.495\linewidth]{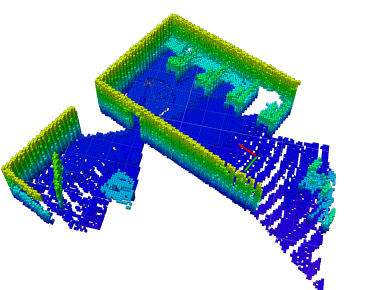}
    };
    \node[fill=white,draw=black,text=black, anchor=south west] at (a.south west) {\footnotesize (a)};

    \node[anchor=north west,inner sep=0] (b) at (4.35cm, 0cm)
    {
      \includegraphics[width=0.495\linewidth]{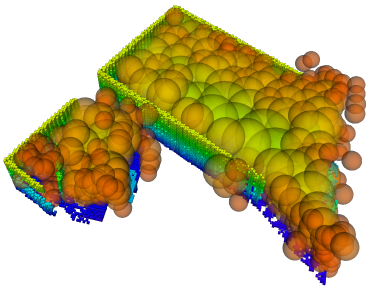}
    };
    \node[fill=white,draw=black,text=black, anchor=south west] at (b.south west) {\footnotesize (b)};

  \end{tikzpicture}
\caption{Map representations using OctoMap (a) and SphereMap with OctoMap (b).}
\label{fig:spheremap}
\end{figure}

An example of such a representation of free space can be seen in Fig. \ref{fig:spheremap}. SphereMap effectively reduces the computational complexity of path planning and obstacle avoidance.

The SphereMap-based accessibility check is computationally efficient enough to operate in real time, even for a large number of \acp{POI}, thanks to precomputed paths. However, in practice, we aim to minimize the number of planned paths to reduce computational overhead. This is achieved by identifying the most interesting \acp{POI} by considering other criteria, as described in the following section.

\subsection{\acf{MRTA}}
\label{section:alloc}

The primary constraint in choosing algorithms for our mission is the necessity of real-time execution on board one of the \acp{UAV} in a dynamic environment. 
We compared two approaches: a simple greedy algorithm based on Euclidean distance and direction from the \ac{UAV} and a more complex algorithm that formulates a \ac{MCF} problem.

We used the greedy approach to
verify the functionality of the system and provide a baseline for comparison.
To minimize the time required to change the heading of the \ac{sUAV}, the greedy approach prefers points located in the same direction as the \ac{sUAV} is facing. For \ac{pUAV}, the heading change is not necessary, thanks to the $360^\circ$ horizontal \ac{FOV} of the 3D \ac{LiDAR}.
We propose to compute the costs $c_{i_P}$ and $c_{i_S}$ for each \ac{POI} $\boldsymbol{g}_i \in \mathcal{G}$ as
\begin{align}
    c_{i_P} &= \norm{\boldsymbol{x}_P - \boldsymbol{g}_i}_2 \label{eq:cost-p} \\
    c_{i_S} &= \alpha \norm{\boldsymbol{x}_S - \boldsymbol{g}_i}_2 + \beta \lvert \phi_S - \theta_i \rvert, \label{eq:cost-s}
\end{align}
where $\alpha , \beta \in \mathbb{R}$ are predefined parameters.

The \acp{UAV} are assigned priority in goal selection to consider the different \ac{UAV} sizes. The higher priority is assigned to \ac{pUAV} as it fits only in spacious areas.
The algorithm identifies the closest accessible point $\boldsymbol{g}_{P_n}$ for the \ac{pUAV} by planning on SphereMap. Subsequently, it finds the first accessible point $\boldsymbol{g}_{S_n}$ for the \ac{sUAV} using the same method, excluding the goal $\boldsymbol{g}_{P_n}$ chosen for the \ac{pUAV}. If no accessible point is found, the \ac{UAV} stays at its current position. 

The advantages of such an approach include speed, low computational complexity, simplicity, and memory efficiency. However, its drawbacks lie in treating the UAVs independently rather than as a unified system. Furthermore, it does not consider path length, which can lead to high travel time despite proximity to the selected point.

To address these issues, we propose to represent
the \acp{UAV} and the \acp{POI} as a bipartite graph $G = (U, V, E)$, where $\boldsymbol{x}_P, \boldsymbol{x}_S \in U$ are nodes represented by \acp{UAV}' position, $\boldsymbol{g}_i \in V$ is a node which stands for corresponding \ac{POI} $i$, and $(\boldsymbol{x}, \boldsymbol{g}, c) \in E$ are edges defined by three values: starting node, end node, and cost $c$. 

Thanks to the graph properties, we can formulate a flow problem.
That requires a definition of the following terms:
\begin{itemize}
    \item \textbf{Source Node} $S$ is a single node that represents the start of the flow.
    \item \textbf{Sink Node} $T$ is a single node representing the end of the flow.
    \item \textbf{Arcs} $\mathcal{A}$ from a UAV to a \ac{POI} exists if the UAV can safely visit the \ac{POI}. The cost $c$ of an arc is the weight of the corresponding edge in the bipartite graph.
    \item \textbf{Capacity} $u$ of an arc from a UAV to a \ac{POI} is one if the UAV can visit the \ac{POI}. Otherwise, the capacity is 0 (in our case, these arcs are not created).
    \item \textbf{Balance} $b$ refers to the conservation of flow at each node in the network. Each feasible solution must satisfy the fact that the amount of flow leaving minus the amount of flow entering the node must be equal to the node's balance. 
\end{itemize}

Let $\mathcal{N}$ be the set of all nodes, $f(a) \in \mathbb{R}_0^+$ the flow on arc $a \in \mathcal{A}$, $\delta^+(n)$ the set of arcs leaving node $n \in \mathcal{N}$, $\delta^-(n)$ the set of arcs entering node $n \in \mathcal{N}$, $c(a)$ the cost per unit of flow on arc $a \in \mathcal{A}$, $u(a)$ the capacity of arc $a \in \mathcal{A}$, and $b(n)$ the balance of node $n \in \mathcal{N}$. The \ac{MCF} problem with zero lower capacity bound can be formulated as
\begin{align}
    \min_{a \in \mathcal{A}}\ &c(a) \cdot f(a) \\
    \mathrm{s.t.} &\sum_{ \forall a \in \delta^+(n)} f(a) - \sum_{ \forall a \in \delta^-(n)} f(a) = b(n), \ \forall n \in \mathcal{N}, \\
    &0 \le f(a) \le u(a), \ \forall a \in \mathcal{A}.
\end{align}

\begin{figure}
    \centering
    \includegraphics[width=\linewidth]{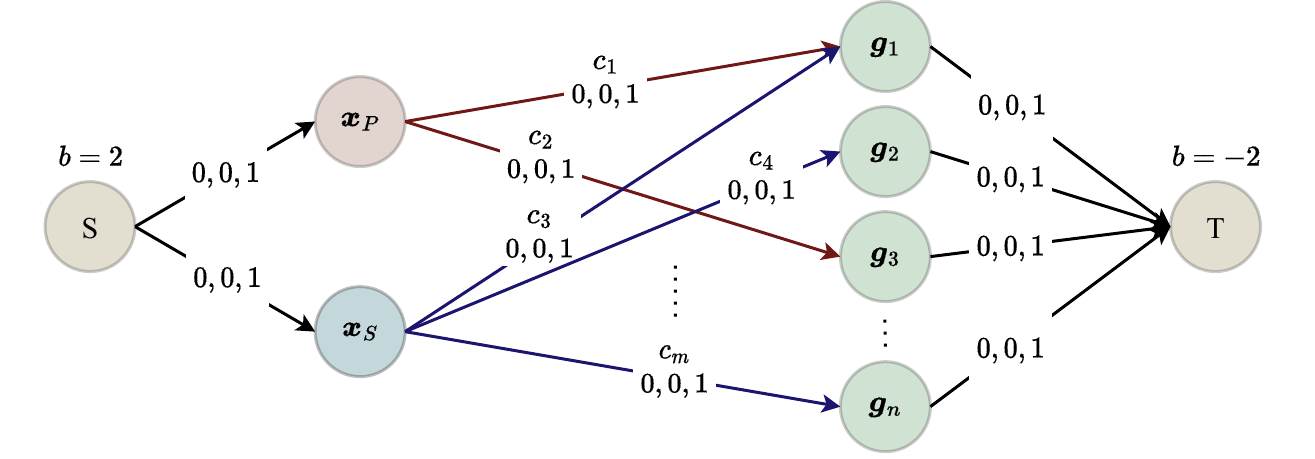}
    \caption[Representation of the assignment problem as a minimum cost flow problem]{Representation of the assignment problem as a minimum cost flow problem. The notation of edge labels $0, 0, 1$ represent capacity's lower bound, current value, and upper bound $u$, respectively.}
    \label{fig:mcf-graph}
\end{figure}

The costs from the source node to each \ac{UAV} $\boldsymbol{x}_i$ and from each \ac{POI} $\boldsymbol{g}_i$ to the sink node are set to 0. These arcs exist only to transfer a matching problem to a flow problem and are irrelevant to the final solution. The balance for each node except source $S$ and sink $T$ is set to 0 (see Fig. \ref{fig:mcf-graph} for an example).

Goal selection based on solving the \ac{MCF} problem is described in Alg. \ref{alg:assign}. The algorithm creates priority queues, $\mathcal{Q}_P$ and $\mathcal{Q}_S$, to manage the \acp{POI}' order based on the distance from the \ac{pUAV} and the combined metric of distance and heading change from the \ac{sUAV}. These computations are identical to Eq. \ref{eq:cost-p} and \ref{eq:cost-s}. The assignment of the next goals involves iteratively selecting \acp{POI} $\boldsymbol{g}$ from these priority queues, computing paths using SphereMap planning, and assessing the feasibility of the paths, given the \ac{UAV} sizes $s_P$ and $s_S$.

This algorithm proceeds to construct arcs, $\mathcal{A}_P$ and $\mathcal{A}_S$, connecting the \acp{UAV} with accessible \acp{POI}, considering their associated costs and path length. Utilizing the priority queues $\mathcal{Q}_P$ and $\mathcal{Q}_S$ to examine the \acp{POI} in the specific order based on Eq. \ref{eq:cost-p} and \ref{eq:cost-s} minimizes the computational complexity. Parameter $N$ is set, which indicates a sufficient number of arcs for each \ac{UAV}, meaning that when there is already $N$ accessible \acp{POI} found, no additional arcs for that \ac{UAV} are created. 

To ensure that the problem can be solved even if $\mathcal{A}_P$ or $\mathcal{A}_S$ is empty, special arcs $(\boldsymbol{x}_P, \boldsymbol{g}_{x_P}, c_{x_P})$ and $(\boldsymbol{x}_S, \boldsymbol{g}_{x_S}, c_{x_s})$, with high costs $c_{x_P}$ and $c_{x_S}$, are created. Nodes $\boldsymbol{g}_{x_P}$ and $\boldsymbol{g}_{x_S}$, added to the \acp{POI}' set, represent current $\boldsymbol{x}_P$ and $\boldsymbol{x}_S$ positions. So, if no accessible \acp{POI} are found, the UAV remains on its current position.

Then, the \ac{MCF} problem is formulated. The nodes $\mathcal{N}$ include: source $S$, sink $T$, the current positions of the \acp{UAV} $\boldsymbol{x}_P$ and $\boldsymbol{x}_S$, the set of \acp{POI} $\mathcal{G}$, and the added points $\boldsymbol{g}_{x_P}$ and $\boldsymbol{g}_{x_S}$. The set of all arcs $\mathcal{A}$ contains arcs $\mathcal{A}_{source}$ from source $S$ to each \ac{UAV}, arcs $\mathcal{A}_{sink}$ from each \ac{POI} $\boldsymbol{g}_{i} \in \mathcal{G}$ to sink $T$, and computed arcs $\mathcal{A}_P$ and $\mathcal{A}_S$. The goals $\boldsymbol{g}_{P_n}$ and $\boldsymbol{g}_{S_n}$ are obtained by solving the \ac{MCF} problem and selecting the \acp{POI} connected with each \ac{UAV} position $\boldsymbol{x}_P$, $\boldsymbol{x}_S$ in the resulting graph with arc capacity 1.

\begin{algorithm}
\caption{Task allocation - assignment problem}
\label{alg:assign}
\KwIn{set of \acp{POI} $\mathcal{G}$, \ac{pUAV} position $\boldsymbol{x}_P$, \ac{sUAV} pose $(\boldsymbol{x_S}, \phi_S)$, source node $S$, sink node $T$, arcs from source node $\mathcal{A}_{source}$, arcs to sink node $\mathcal{A}_{sink}$}
\KwOut{\ac{pUAV} next goal $\boldsymbol{g}_{P_n}$, \ac{sUAV} next goal $\boldsymbol{g}_{S_n}$}
\Parameters{\ac{pUAV} size $s_P$, \ac{sUAV} size $s_S$, cost $c_{x_P}$, cost $c_{x_S}$, number of maximum arcs $N$ for each \ac{UAV}}
\BlankLine
$\boldsymbol{g}_{x_P} \gets \boldsymbol{x}_P$, $\boldsymbol{g}_{x_S} \gets \boldsymbol{x}_S$ \\
$\mathcal{Q}_P \gets$ \textsc{priority\_queue}($\mathcal{G}, \boldsymbol{x}_P$) 
\Comment{based on distance from \ac{pUAV}} \\
$\mathcal{Q}_S \gets$ \textsc{priority\_queue}($\mathcal{G}, \boldsymbol{x}_S, \phi_S$) 
\Comment{based on distance and heading change from \ac{sUAV}} \\

$\mathcal{A_P} \gets \emptyset$ \Comment{arcs from \ac{pUAV}} \\
$i \gets 0$ \\
\While{$i < N$ and $\mathcal{Q}_P \ne \emptyset$}{
    $c, \boldsymbol{g} \gets \mathcal{Q}_P$.\textsc{pop}() \Comment{cost and \ac{POI}}\\
    $\mathcal{P} \gets$ \textsc{find\_path}$(\boldsymbol{x}_P, \boldsymbol{g}, s_P)$ \Comment{SphereMap}\\
    \If{$\mathcal{P} \ne \emptyset$} {
        $c \gets c +$\textsc{length}$(\mathcal{P})$ \Comment{number of waypoints} \\
        $\mathcal{A}_P \gets \mathcal{A}_P \cup \{(\boldsymbol{x}_P, \boldsymbol{g}, c)\}$ \\
        $i \gets i + 1$ \\
    }
}
$\mathcal{A}_P \gets \mathcal{A}_P \cup \{(\boldsymbol{x}_P, \boldsymbol{g}_{x_P}, c_{x_P})\}$ \Comment{to secure feasibility} \\
repeat lines 4-15 for \ac{sUAV} with $\mathcal{Q}_S, \boldsymbol{x}_S, s_S, \boldsymbol{g}_{x_S}, c_{x_S}$, and $\mathcal{A}_S$ \\
$\mathcal{N} \gets \left\{S, T, \boldsymbol{x}_P, \boldsymbol{x}_S\right\} \cup \mathcal{G} \cup \{\boldsymbol{g}_{x_P},\boldsymbol{g}_{x_S}\}$ \\
$\mathcal{A} \gets \mathcal{A}_{source} \cup \mathcal{A}_{sink} \cup \mathcal{A}_P \cup \mathcal{A}_S$ \\
$\boldsymbol{g}_{P_n}, \boldsymbol{g}_{S_n} \gets$ \textsc{solve\_min\_cost\_flow}$(\mathcal{N}, \mathcal{A})$ 
\end{algorithm}

\subsection{Path Planning}
\label{section:planning}

The proposed path planning algorithm operates
in two states: \textsc{monitoring} and \textsc{planning}. In the \textsc{monitoring} state, the algorithm checks if either \ac{UAV} $u \in \mathcal{U}$, where $\mathcal{U} = \left\{\textrm{\acs{pUAV}, \acs{sUAV}}\right\}$, has reached its current goal $\boldsymbol{g}_{u_c}$, creating a set of \acp{UAV} waiting for new path $\mathcal{W}$ and updating the set of already visited points $\mathcal{V}$. When at least one \ac{UAV} is waiting for a new action, the algorithm transitions to the \textsc{planning} state.

The \textsc{planning} stage is triggered by a map $\mathcal{M}$ update to ensure that newly explored areas are also searched for \acp{POI}. Then, paths for each \ac{UAV} $w \in \mathcal{W}$ are computed using the SphereMap. This approach is aligned with the priorities of fast computation and safety since the \ac{sUAV} is designed to fly through narrow spaces. We did not use the precomputed paths during the planning process to obtain a smoother path.

The computed paths are transformed into the \acp{UAV}' local frames. For \ac{pUAV}, the transformation matrix $\prescript{\mathrm{P}}{\mathrm{G}}{\boldsymbol{T}}$ is identity, and for \ac{sUAV}, the matrix $\prescript{\mathrm{S}}{\mathrm{G}}{\boldsymbol{T}}$ is continuously updated by the relative localization.

Each UAV is then instructed to follow its respective path, and the algorithm returns to the \textsc{monitoring} state.

\subsection{Obstacle Avoidance}
\label{section:safety}

We assume that the environment does not contain dynamic obstacles, therefore only static obstacles and the possible collision of the two \acp{UAV} are considered. The static obstacles are handled by the planning process, and collision avoidance between \acp{UAV} is secured by creating 
three safety zones based on the \acp{UAV}' distance. The first one, bounded by the critical distance $d_C$ parameter, marks the area where the \acp{UAV} are too close to each other, and an avoidance maneuver is necessary. The maneuver takes advantage of the small size of the \ac{sUAV}. The \ac{pUAV} stops and is stationary the whole time. 
A new goal position $\boldsymbol{g}_S$ is computed and assigned to the \ac{sUAV} (the value of the z-axis stays the same as the current altitude):
\begin{align}
    \boldsymbol{u} &= \frac{\boldsymbol{x}_S - \boldsymbol{x}_P}{\norm{\boldsymbol{x}_S - \boldsymbol{x}_P}_2} \\
    \boldsymbol{g}_S &= \boldsymbol{x}_S + \boldsymbol{u} \\
    \boldsymbol{g}_{S_z} &= \boldsymbol{x}_{S_z} \\
    \theta_S &= \phi_S
\end{align}

The path $\mathcal{P}_S$ to the goal $\boldsymbol{g}_S$ is planned by an A$^*$-based planner \cite{subt-planner}, 
on global map $\mathcal{M}$, ensuring that it is feasible, short, and safe. A path is found even if the goal $\boldsymbol{g}_S$ is not accessible, but there is a reachable point in its proximity. This path is transformed to \ac{sUAV} \ac{VIO} frame $\mathrm{S}$ using transformation matrix $\prescript{\mathrm{S}}{\mathrm{G}}{\boldsymbol{T}}$ and sent to the control pipeline. When the \ac{sUAV} reaches its goal, the whole process is repeated until the distance between the \acp{UAV} is bigger than the critical distance $d_C$. 

In the second zone, when the distance is smaller than the safety distance $d_S$ parameter but bigger than $d_C$, the safety of \ac{pUAV} is prioritized, as it is equipped with more precise and more expensive sensors. The \ac{pUAV} stops its action and waits until the \ac{sUAV}, which continues in its current action, is far enough and the safety condition is satisfied:
\begin{align}
    \norm{\boldsymbol{x}_P - \boldsymbol{x}_S}_2 \ge d_S
\end{align}

The last zone, where the distance is bigger than the safety distance $d_S$, is marked as safe, and the \acp{UAV} explore the space as described in previous sections.


\section{\textsc{Experimental Verification}}
We used two types of \acp{UAV} (see Fig. \ref{fig:uav-paltform}) that are controlled by the Pixhawk 4 Flight Controller with built-in \acp{IMU}. The \ac{pUAV} is based on the Holybro X500 frame and has a size of 0.7 by 0.7 meters, including propellers. It features a powerful onboard computer, the Intel NUC 10iFNH, with a Wi-Fi module for data transmission, and carries a 3D \ac{LiDAR} Ouster OS0-128 Rev D, with a $360^\circ$ horizontal and $90^\circ$ vertical \ac{FOV}, and range of 40 meters.

The \ac{sUAV}, built around the DJI F330 frame with compact dimensions of approximately 0.45 by 0.45 meters, propellers included, is equipped with Intel NUC 10iFNH, the RealSense T265 tracking camera, and the RealSense D435 depth camera. The RealSense T265 incorporates two fisheye lens sensors with a $173^\circ$ diagonal \ac{FOV}, and an \ac{IMU}. The RealSense D435 stereo-depth features \ac{IR} sensors with an $87^\circ$ by $58^\circ$ \ac{FOV} and an \ac{RGB} sensor with a $69^\circ$ by $42^\circ$ \ac{FOV}, alongside an \ac{IR} projector with a range of 5 meters. For details of the platform setup, see \cite{hardware-decs2}.

The software is based on Ubuntu 20.04 and \ac{ROS} 1 Noetic, and the \acp{UAV} run on the \ac{MRS system} \cite{bib:mrs-system}, which allows for easy switching between simulation and real-world scenarios. Self-localization is achieved through \ac{SLAM} algorithm \ac{LOAM} \cite{loam}, specifically \ac{A-LOAM} for the \ac{pUAV}, and the OpenVINS algorithm \cite{openvins} offers a state-of-the-art, filter-based, \ac{VIO} for the \ac{sUAV}.

\subsection{Simulations}
\begin{figure}
\centering
\begin{tikzpicture}
    \node[anchor=north west,inner sep=0] (a) at (0, 0)
    {
      \includegraphics[width=.495\linewidth, trim={1.5mm 1.5mm 1.5mm 1.5mm},clip]{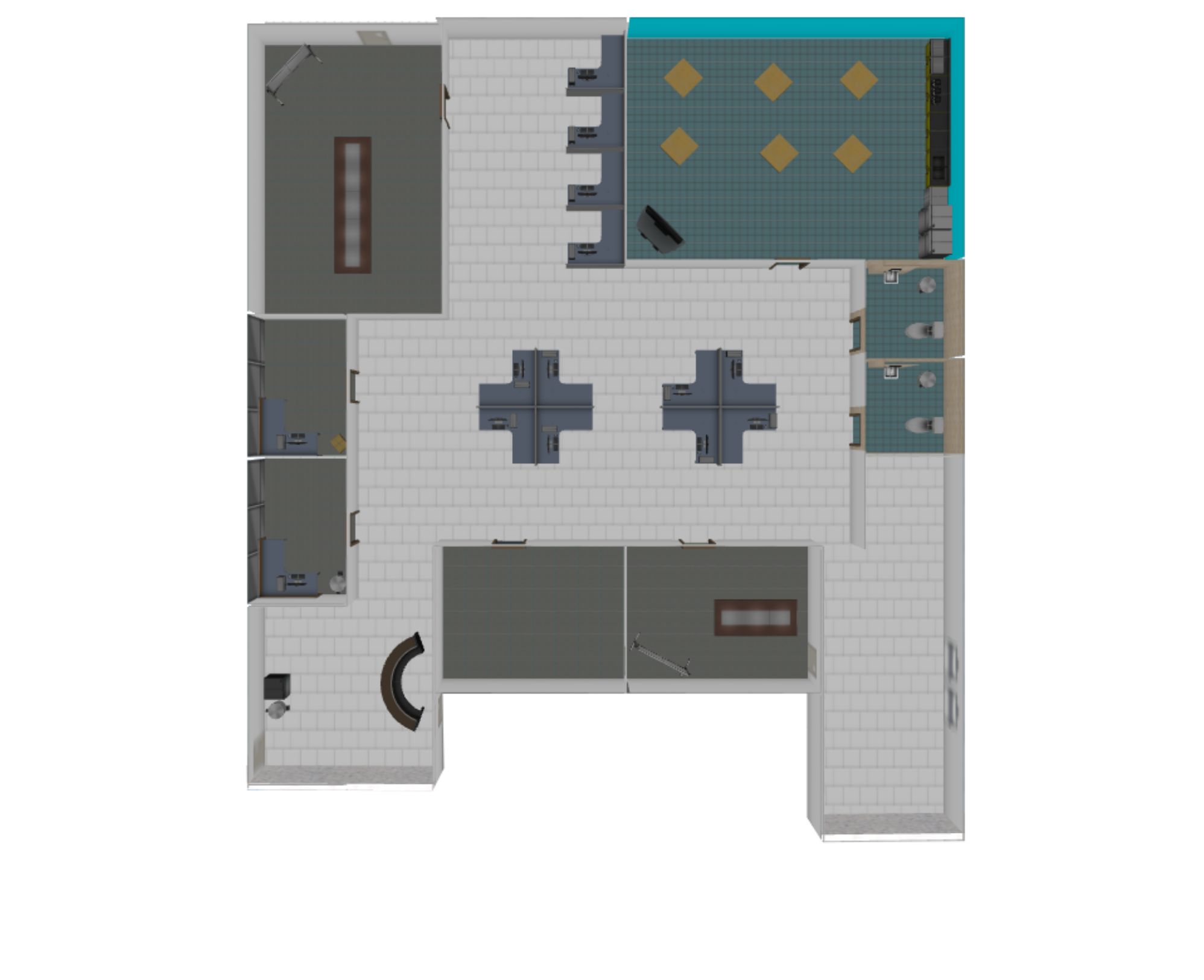}
    };
    \node[fill=white,draw=black,text=black, anchor=south west] at (a.south west) {\footnotesize (a)};

    \node[anchor=north west,inner sep=0] (b) at (4.35cm, 0cm)
    {
      \includegraphics[width=.495\linewidth, trim={1.5mm 1.5mm 1.5mm 1.5mm},clip]{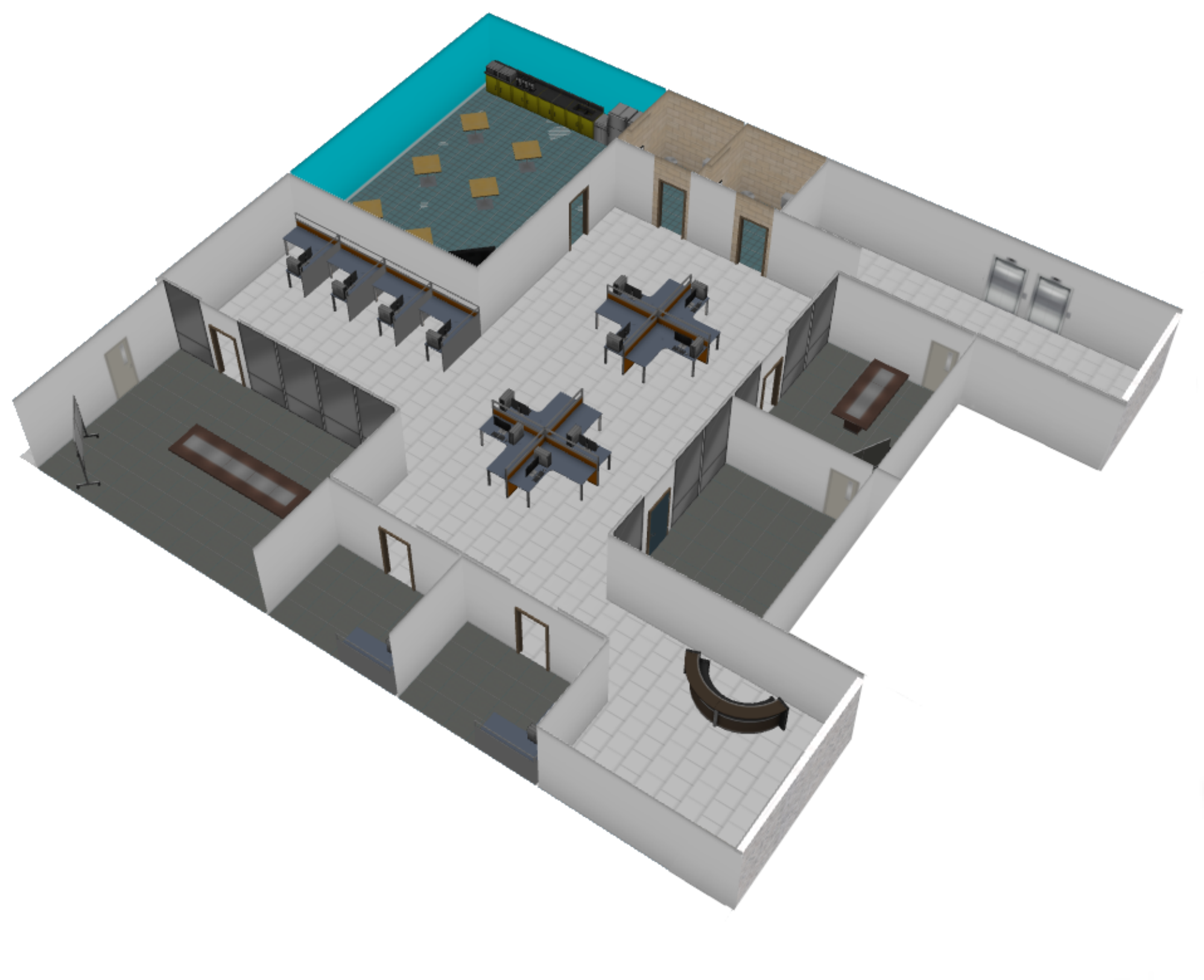}
    };
    \node[fill=white,draw=black,text=black, anchor=south west] at (b.south west) {\footnotesize (b)};

  \end{tikzpicture}
\caption{Gazebo model of an office environment used for the experiments: top view (a), side view (b).}
\label{fig:office-gazebo}
\end{figure}

The \acp{UAV} were tasked to explore an office environment of $23 \times 23 \times 3$ meters created in the Gazebo robotic simulator, visualized in Fig. \ref{fig:office-gazebo}. A spacious common area and multiple smaller rooms accessible through open doors, $0.9$ meters wide, where only the \ac{sUAV} can safely fit, were used to verify the cooperation between heterogeneous \acp{UAV}. Tab. \ref{tab:sim-param} provides a list of parameters used for the simulations.

To evaluate the performance of the proposed algorithms in isolation from localization errors and to reduce the computational demands of the simulation, ground-truth data were used for self-localization of both \acp{UAV} and relative localization between the \acp{UAV}. 

\begin{table}
    \centering
    \caption{Parameters for the simulation experiments}
    \begin{tabular*}{\linewidth}{l@{\extracolsep{\fill}}cc}
        \hline
        \textbf{Parameter} & \textbf{Notation} & \textbf{Value} \\ \hline
        OctoMap voxel size & $ \mathcal{M}_{res}$ & 0.1 m \\
        \ac{pUAV} size (radius) & $s_P$ & 0.45 m \\
        \ac{sUAV} size (radius) & $s_S$ & 0.25 m \\
        minimum safety distance & $d_S$ & $2.5$ m \\
        minimum critical distance & $d_C$ & $2.0$ m \\
        minimum radius of spheres & $r_{\mathrm{sph}}$ & 0.35 m \\
        frontier detection rate & $F_{\mathrm{front}}$ & 0.5 Hz \\
        path planning rate & $F_{\mathrm{path}}$ & 2 Hz \\
        collision avoidance rate & $F_{\mathrm{coll}}$ & 10 Hz \\
        \hline
    \end{tabular*}
    \label{tab:sim-param}
\end{table}

\begin{figure}
\centering
\begin{tikzpicture}
    \node[anchor=south west,inner sep=0] (image) at (0,0) {\includegraphics[width=\linewidth, angle=-90,origin=c, trim={1mm 1mm 1mm 1mm},clip]{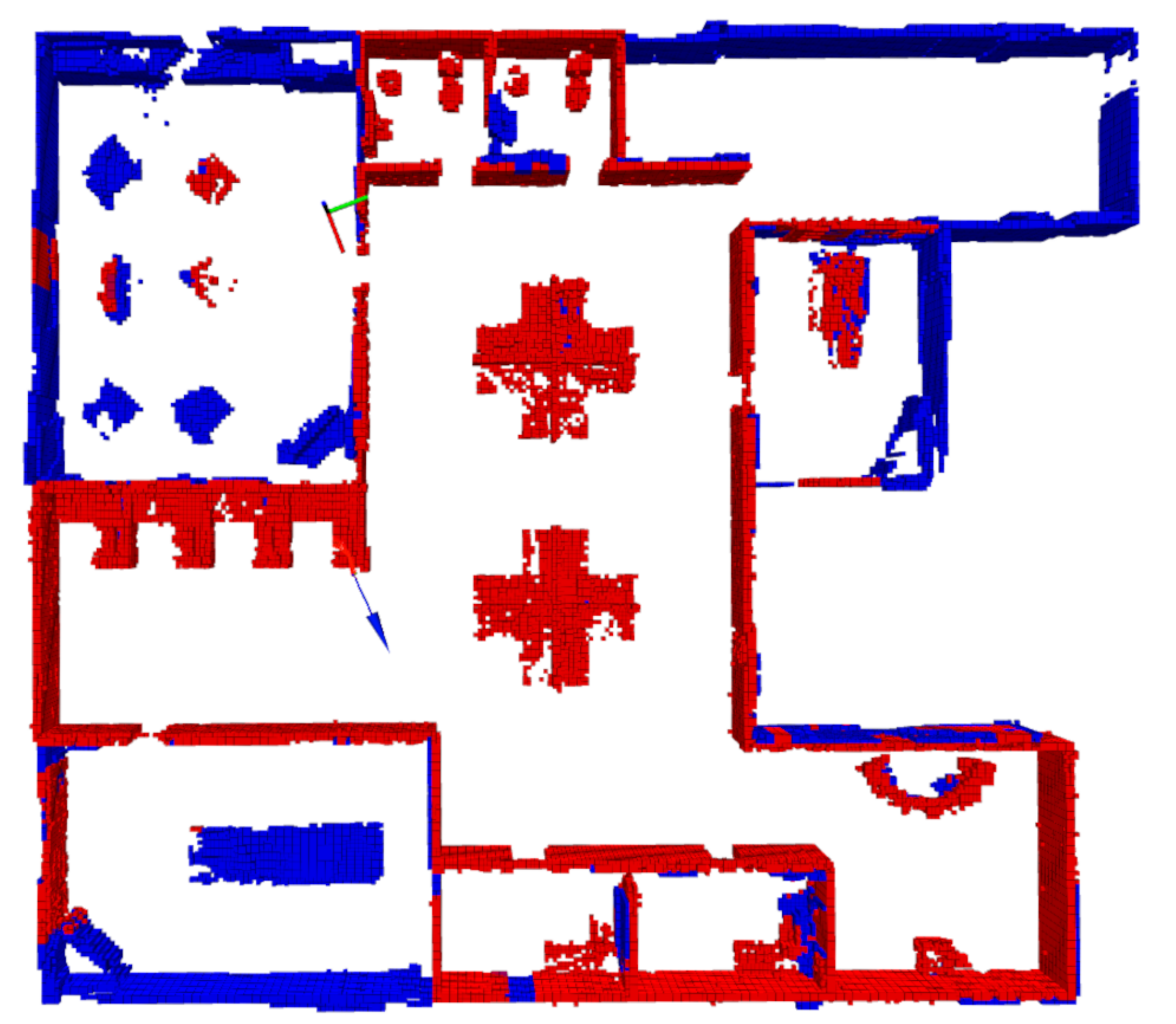}};
    \node[draw, rounded corners] at (5.3, 6.5) {sUAV};
    \node[draw, rounded corners] at (2.7, 6.3) {pUAV};
\end{tikzpicture}
\caption[Occupancy map of the office space]{Occupancy map of the office space at the end of the 4-minute mission. The red color represents an area scanned by the \acs{pUAV} using \acs{LiDAR}, and the blue color represents space explored only by \acs{sUAV}.}
\label{fig:office-map}
\end{figure}

Fig. \ref{fig:office-map} shows the final occupancy map. It was obtained as the result of an exploration mission that lasted 4 minutes using the greedy approach to solve the task allocation problem. The common, spacious area was explored by the \ac{pUAV}, whereas the smaller rooms, accessible only through the doors, were explored only by the \ac{sUAV}. For clarity, the floor and ceiling were removed from the visualization.

\begin{figure}
\centering
\begin{tikzpicture}
    \node[anchor=north west,inner sep=0] (a) at (0, 0)
    {
      \includegraphics[width=0.495\linewidth]{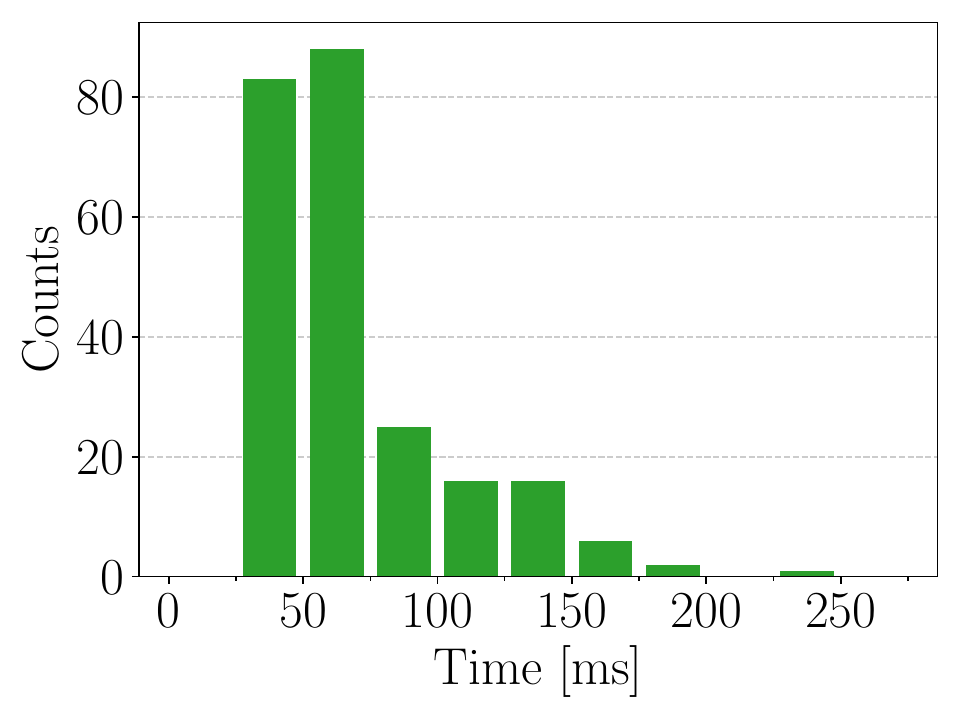}
    };
    \node[fill=white,draw=black,text=black, anchor=south west] at (a.south west) {\footnotesize (a)};

    \node[anchor=north west,inner sep=0] (b) at (4.35cm, 0cm)
    {
      \includegraphics[width=0.495\linewidth]{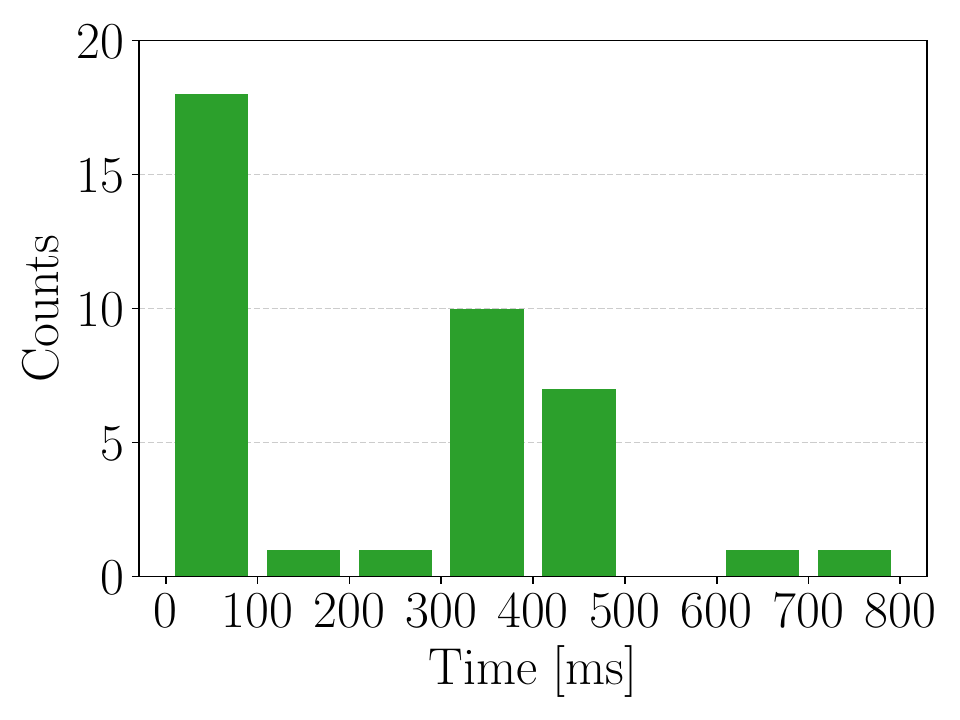}
    };
    \node[fill=white,draw=black,text=black, anchor=south west] at (b.south west) {\footnotesize (b)};

    \node[anchor=north west,inner sep=0] (c) at (0cm, -3.5cm)
    {
      \includegraphics[width=.495\linewidth]{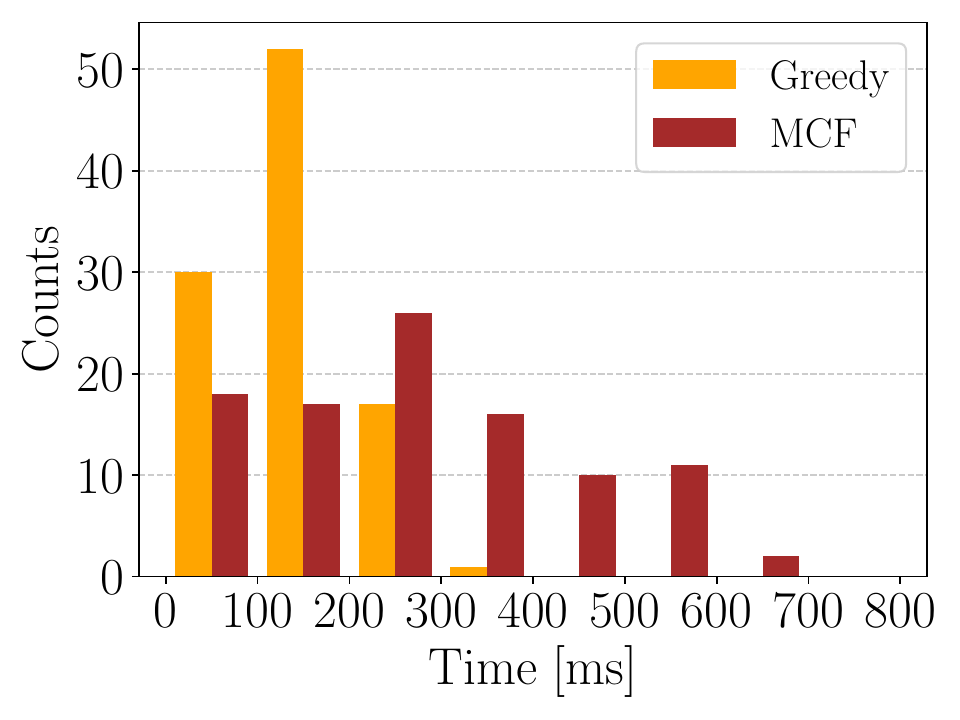}
    };
    \node[fill=white,draw=black,text=black, anchor=south west] at (c.south west) {\footnotesize (c)};

    \node[anchor=north west,inner sep=0] (d) at (4.35cm, -3.5cm)
    {
      \includegraphics[width=.495\linewidth]{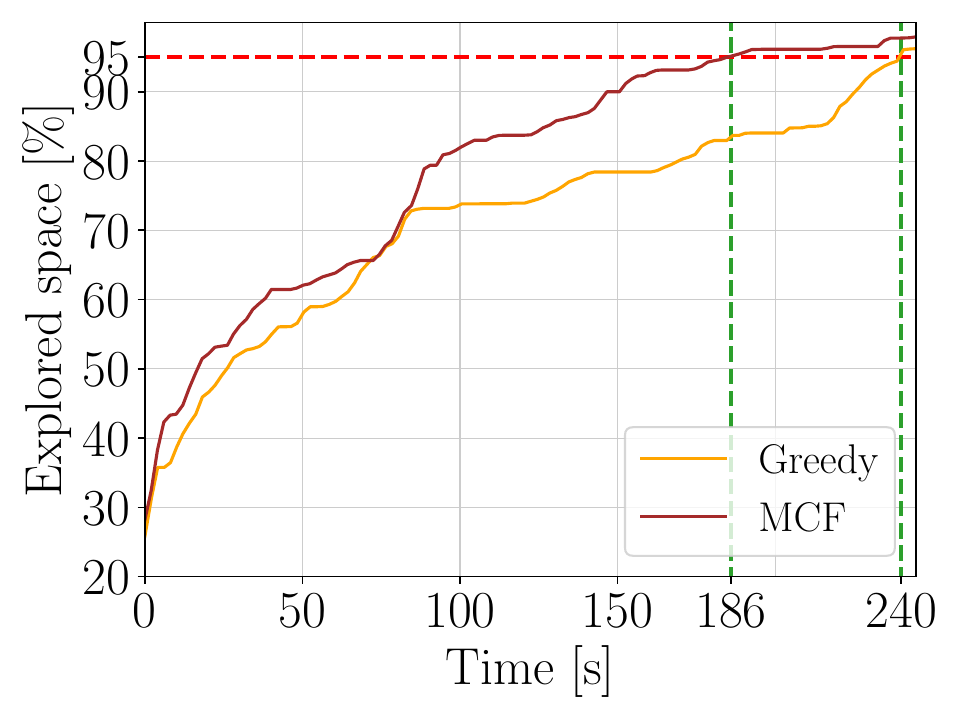}
    };
    \node[fill=white,draw=black,text=black, anchor=south west] at (d.south west) {\footnotesize (d)};
\end{tikzpicture}

\caption{Execution time of the algorithms in the office environment for frontier detection (a), path planning (b), goal assignment (c), and exploration time comparison (d).}
\label{fig:office-hist}
\vspace{-0.8em}
\end{figure}
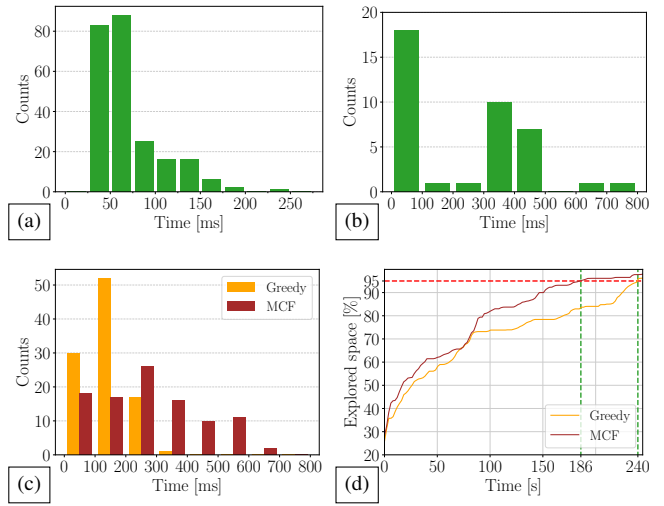

To analyze separate parts of the method, we measured the processing times of frontier detection, goal assignment, and path planning. These values were obtained during the experiments and provide a general idea about the algorithms' complexity. The results are shown in Fig. \ref{fig:office-hist}.

Frontier detection's processing time (see Fig. \ref{fig:office-hist}) depends on the map complexity. In an indoor environment, the processing time is significantly higher at the beginning of the exploration, as most of the leaf nodes to be explored are free. As the exploration progresses, the number of free leaf nodes $l \in \mathcal{M}_{\mathrm{free}}$ is decreasing. The median processing time of frontier detection was $59.64$ ms.

To compare the two algorithms for solving the task allocation problem, experiments with identical exploration parameters (see Table \ref{tab:sim-param}) and identical \acp{UAV}' initial position were executed. The parameter $N$ used in the \ac{MCF} approach was set to $N=5$, meaning the algorithm looks only for the first five accessible \acp{POI} for each \ac{UAV}.

Both approaches successfully explored the environment and created the occupancy map visualized in Fig. \ref{fig:office-map}. However, the \ac{MCF} approach significantly improves the exploration efforts. As shown in Fig. \ref{fig:office-hist}, where the map exploration completion refers to the ratio of the explored volume of the map to the total volume of the map, it took $240.68$ s to explore $95\ \% $ of the space using the greedy approach, whereas the \ac{MCF} algorithm was able to achieve the same result in $186.55$ s. The $95\%$ boundary is marked by a red horizontal dashed line in Fig. \ref{fig:office-hist}, and the green vertical dashed lines correspond to $t_{greedy}$ and $t_{mcf}$. If the greedy approach is considered the baseline solution, the \ac{MCF} results in almost $30\%$ improvement.

However, the drawback of the \ac{MCF} lies in higher execution time. The performance of the two algorithms was evaluated by measuring their respective processing times. The results are shown in Fig. \ref{fig:office-hist} in a histogram of the goal assignment processing time of both algorithms created from 100 samples for each method. Even though the \ac{MCF} approach overall speeds up the exploration, the computation of the global objective is more expensive. The median processing time of the greedy approach was $144.53$ ms and the median processing time for the \ac{MCF} approach was $253.76$ ms. 

The processing time of the path-planning algorithm differs based on the $state$ variable. However, the processing time of the \textsc{monitoring} state is insignificant in comparison to the \textsc{planning} and, therefore, was not measured.
The number of \acp{UAV} waiting for new action has the main effect on the time of the \textsc{planning} state. Fig. \ref{fig:office-hist} shows two peaks in the histogram, corresponding to the situation when either one or both UAVs need new paths.

\subsection{Real-World Experiments}

\begin{figure}
\centering
\begin{tikzpicture}
    \node[anchor=north west,inner sep=0,draw=black] (a) at (0, 0)
    {
      \includegraphics[width=0.495\linewidth, trim={1mm 1mm 1mm 1mm},clip]{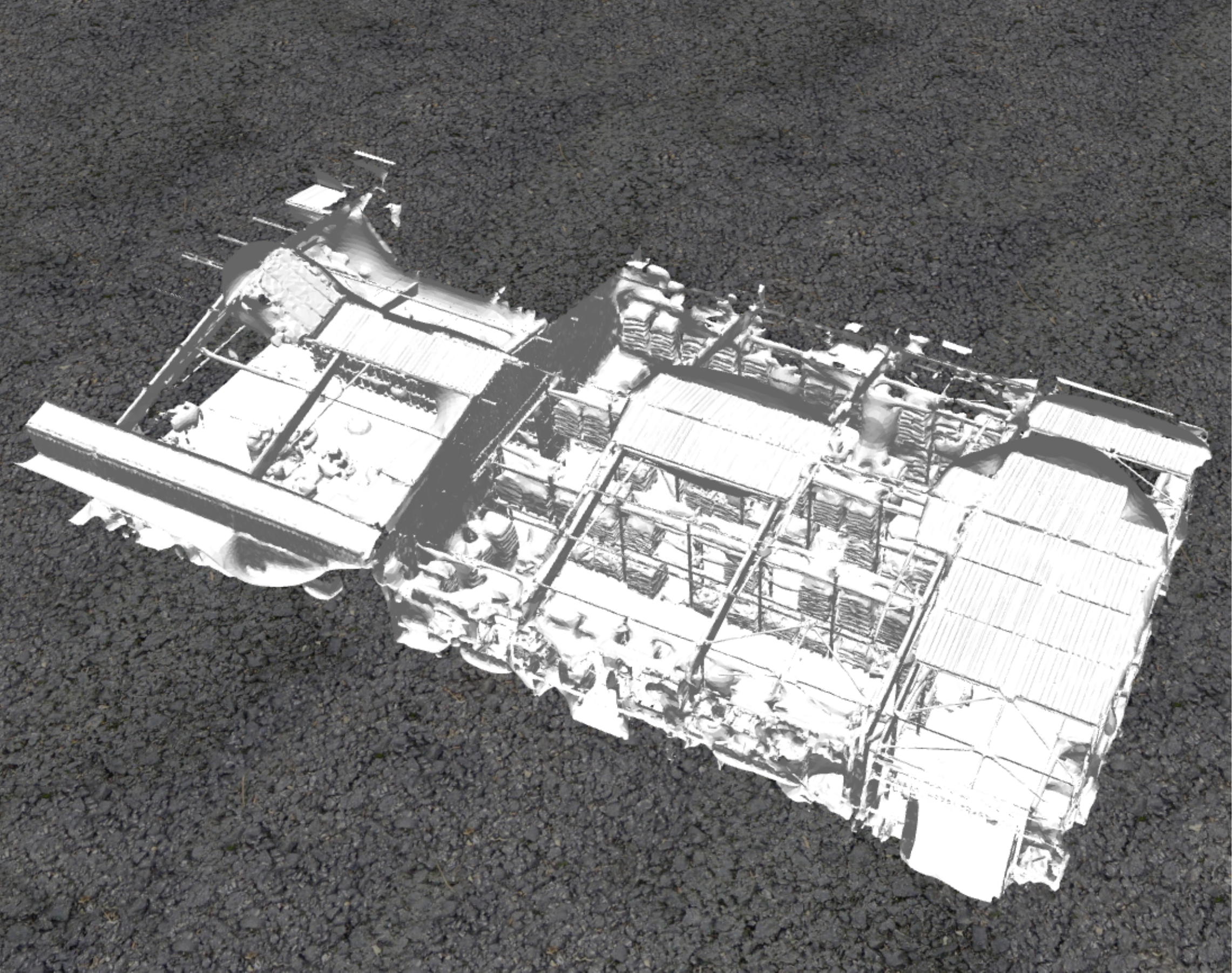}
    };
    \node[fill=white,draw=black,text=black, anchor=south west] at (a.south west) {\footnotesize (a)};

    \node[anchor=north west,inner sep=0,draw=black] (b) at (4.35cm, 0cm)
    {
      \includegraphics[width=0.495\linewidth, trim={1mm 1mm 1mm 1mm},clip]{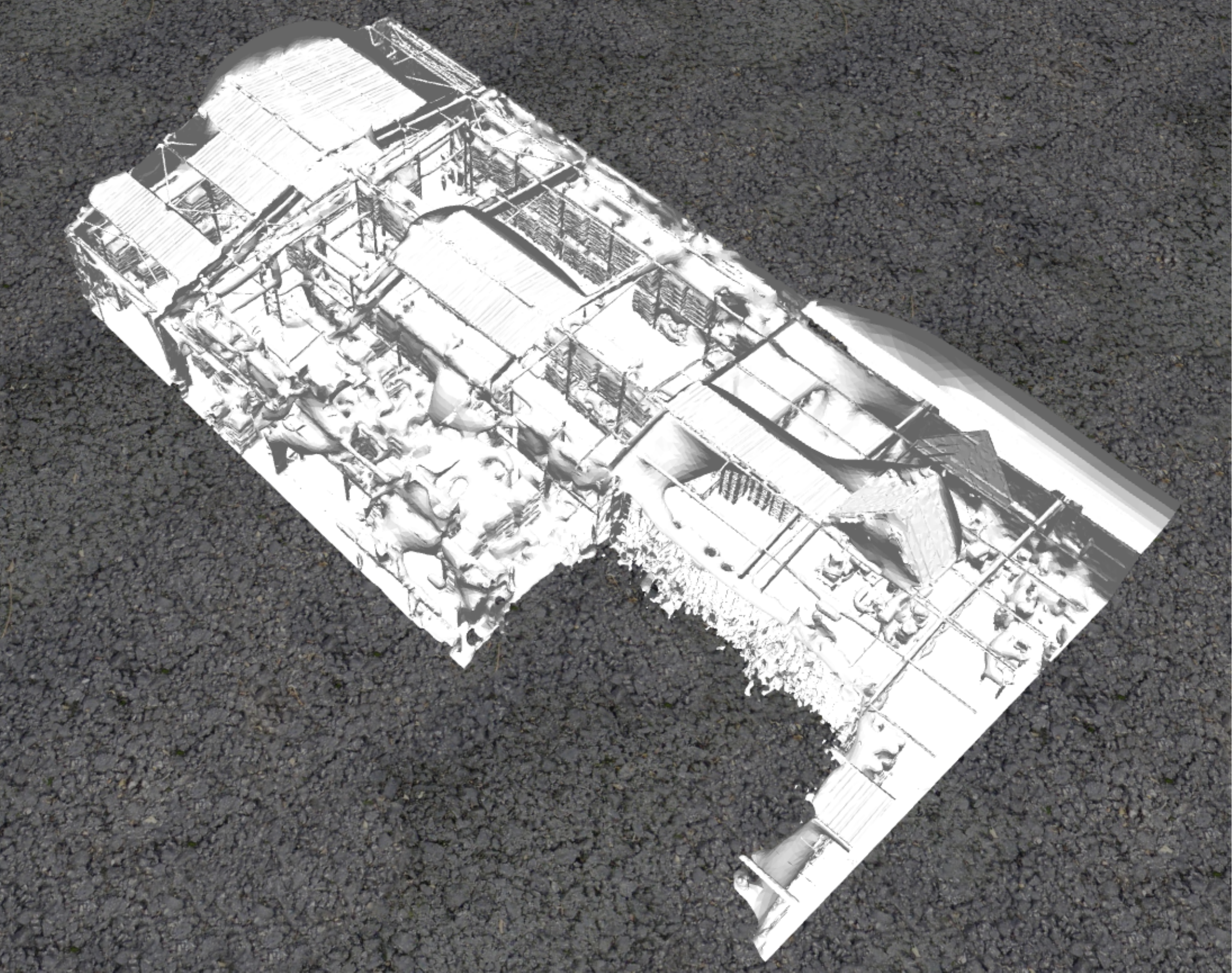}
    };
    \node[fill=white,draw=black,text=black, anchor=south west] at (b.south west) {\footnotesize (b)};

  \end{tikzpicture}
\caption{3D model of a warehouse environment where the real-world experiments were executed: front view (a), back view (b).}
\label{fig:warehouse-gazebo}
\end{figure}

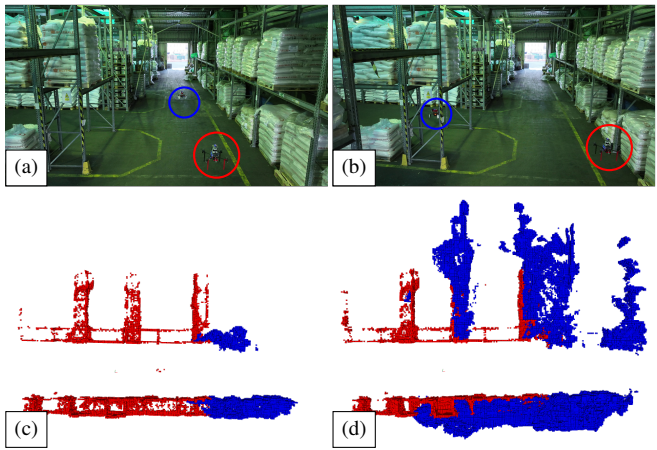
\begin{figure}
\centering
\begin{tikzpicture}
    \node[anchor=north west,inner sep=0,draw=black] (a) at (0, 0)
    {
      \includegraphics[width=.495\linewidth, trim={1mm 1mm 1mm 1mm},clip]{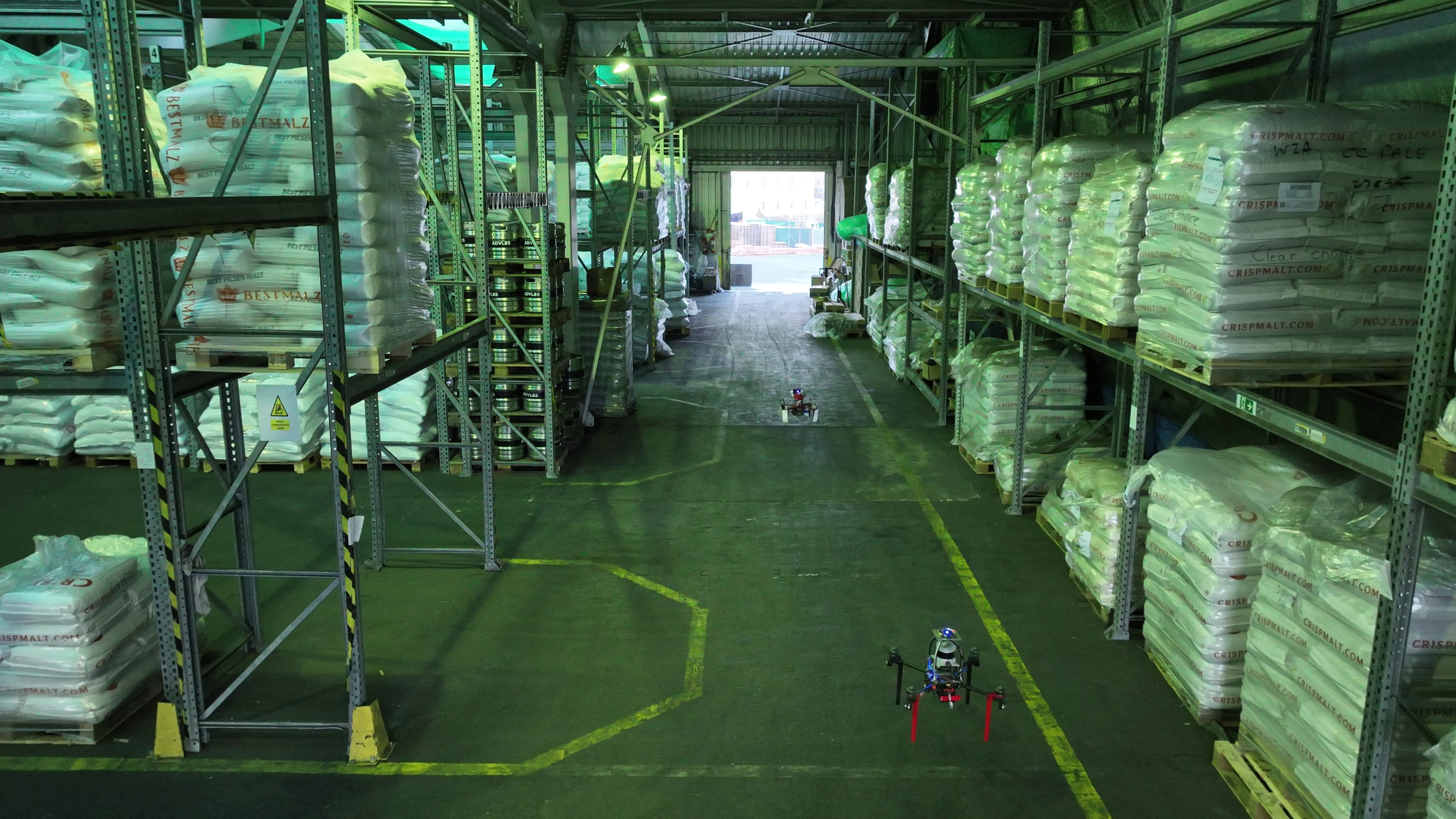}
    };
    \node[fill=white,draw=black,text=black, anchor=south west] at (a.south west) {\footnotesize (a)};
    \draw[thick, color=red] (2.81,-2.0) circle [radius=0.3]; 
    \draw[thick, color=blue] (2.38,-1.3) circle [radius=0.2];

    \node[anchor=north west,inner sep=0,draw=black] (b) at (4.35cm, 0cm)
    {
      \includegraphics[width=.495\linewidth, trim={1mm 1mm 1mm 1mm},clip]{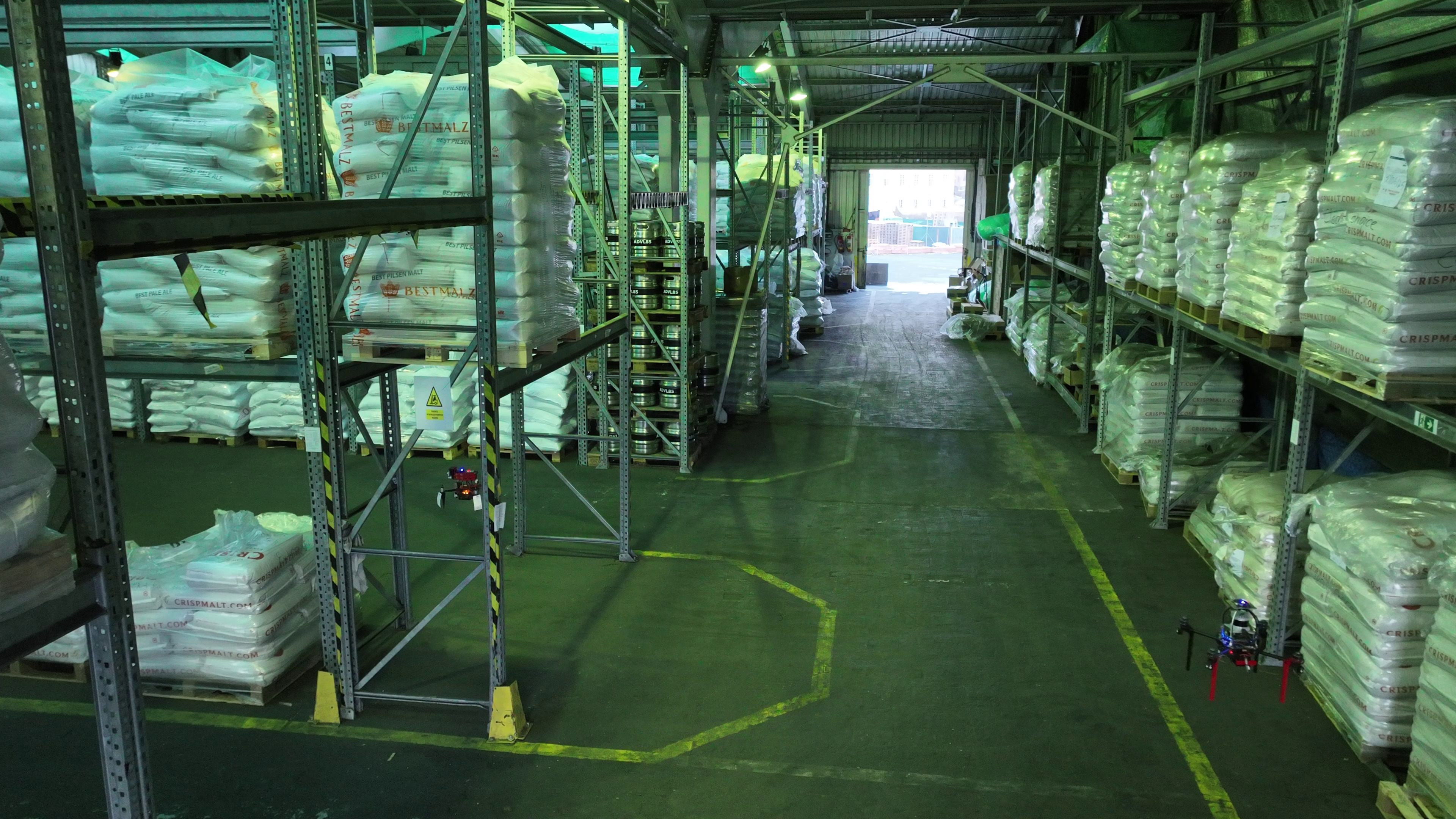}
    };
    \draw[thick, color=red] (8.03,-1.9) circle [radius=0.3]; 
    \draw[thick, color=blue] (5.73,-1.45) circle [radius=0.2];
    \node[fill=white,draw=black,text=black, anchor=south west] at (b.south west) {\footnotesize (b)};

    \node[anchor=north west,inner sep=0] (c) at (0cm, -2.5cm)
    {
      \includegraphics[width=.495\linewidth, trim={1mm 1mm 1mm 1mm},clip]{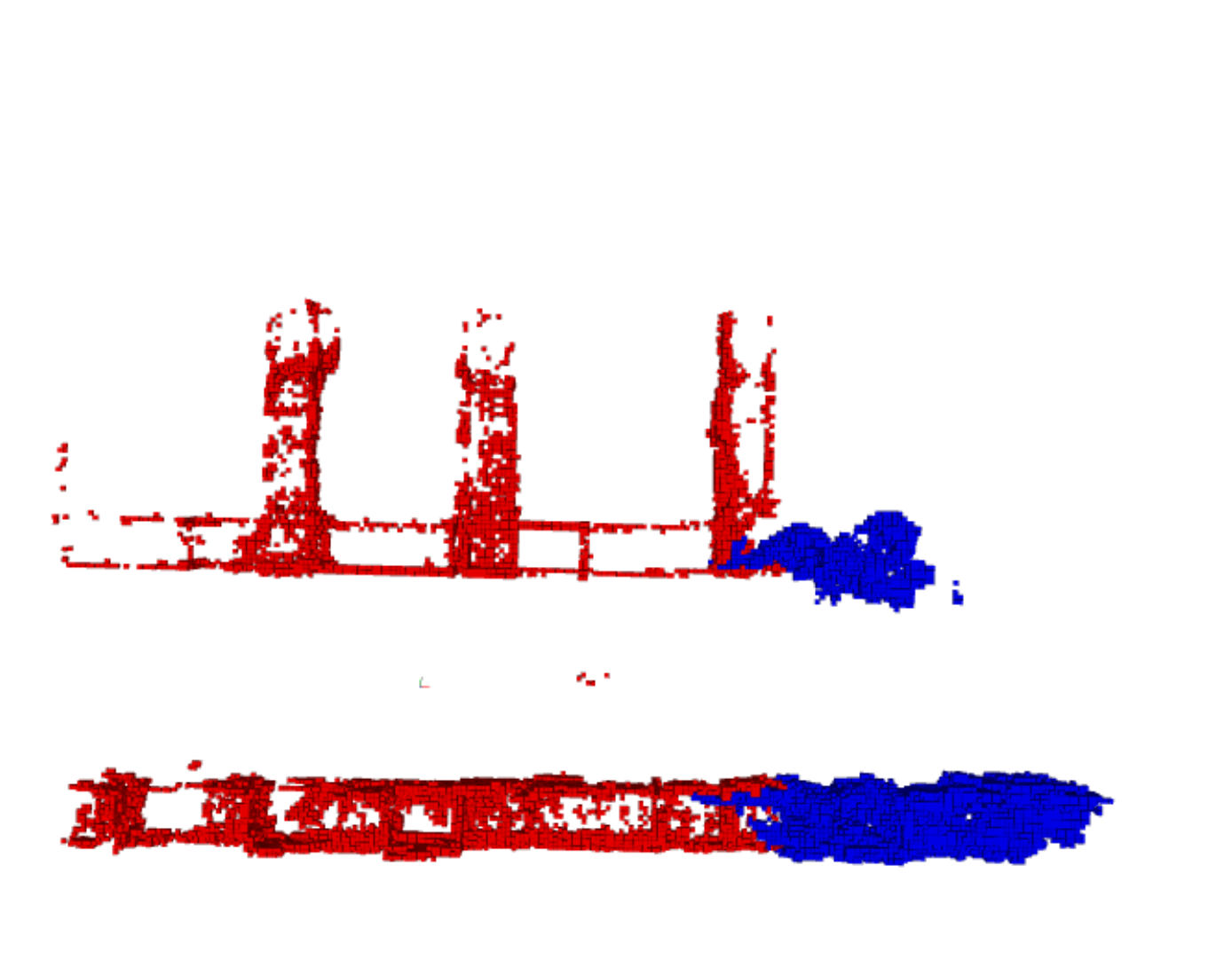}
    };
    \node[fill=white,draw=black,text=black, anchor=south west] at (c.south west) {\footnotesize (c)};

    \node[anchor=north west,inner sep=0] (d) at (4.35cm, -2.5cm)
    {
      \includegraphics[width=.495\linewidth, trim={1mm 1mm 1mm 1mm},clip]{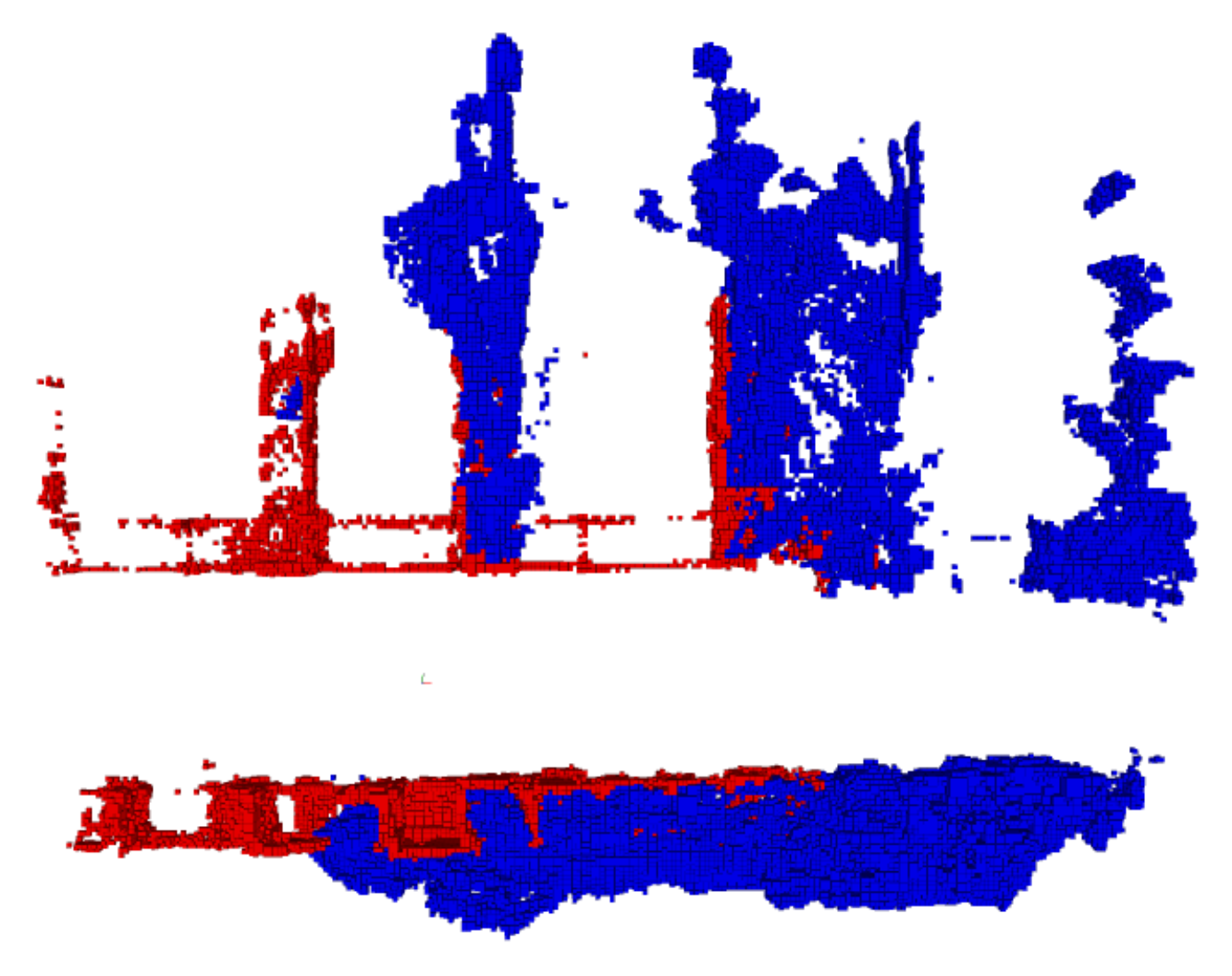}
    };
    \node[fill=white,draw=black,text=black, anchor=south west] at (d.south west) {\footnotesize (d)};

  \end{tikzpicture}

\caption{Photographs from the real-world experiments (a), (b) and 
global occupancy grid at the initial position to show the exploration progress (c) and the final map (d).}
\label{fig:warehouse-real-map}
\end{figure}

We tested the proposed algorithms in a real-world industrial warehouse environment (see Fig. \ref{fig:warehouse-gazebo}) without any external localization system or computational resources. To ensure the safety of the \acp{UAV}, we updated the exploration parameters. The values are listed in Tab. \ref{tab:real-param}.
The final map from one of the successful experiments is shown in Fig. \ref{fig:warehouse-real-map}. This map was obtained after a mission that lasted 4.1 minutes. 

\begin{table}
    \centering
    \caption{Parameters for real-world experiments}
    \begin{tabular*}{\linewidth}{l@{\extracolsep{\fill}}cc}
        \hline
        \textbf{Parameter} & \textbf{Notation} & \textbf{Value} \\ \hline
        OctoMap voxel size & $ \mathcal{M}_{res}$ & 0.1 m \\
        \ac{pUAV} size (radius) & $s_P$ & 1.8 m \\
        \ac{sUAV} size (radius) & $s_S$ & 0.4 m \\
        minimum safety distance & $d_S$ & $4.0$ m \\
        minimum critical distance & $d_C$ & $3.5$ m \\
        minimum radius of spheres & $r_{\mathrm{sph}}$ & 0.9 m \\
        frontier detection rate & $F_{\mathrm{front}}$ & 0.5 Hz \\
        path planning rate & $F_{\mathrm{path}}$ & 2 Hz \\
        collision avoidance rate & $F_{\mathrm{coll}}$ & 10 Hz \\
        \hline
    \end{tabular*}
    \label{tab:real-param}
\end{table}

The experiments provided useful data regarding the real-life execution time of the proposed approach. The environment was complex, so random samples from the frontiers cluster were selected to enlarge the \acp{POI}'s set. 

The assignment challenge was solved using the \ac{MCF}, which outperformed the greedy method in simulation experiments. However, due to security measures, none of the \acp{POI} were reachable by the \ac{pUAV}. This led to the worst-case scenario, where the accessibility of all the \acp{POI} had to be verified by computing and checking their paths at least once. Despite the large number of \acp{POI}, the maximum time required to solve the assignment problem was $1.15$ s. This time was measured during the exploration, which confirms that the algorithm can handle large and complex environments. The median processing time for the frontier detection in this scenario was 178.2 ms, 842.12 ms for the assignment, and 23.49 ms for planning.

During the experiment, we noticed drifts in the estimated pose of the \ac{sUAV} due to localization drift of the \ac{VIO} when the line of sight between the \acp{UAV} was broken for a prolonged period of time. When the estimated pose was not correct, the path followed by the \ac{sUAV} differed from the one planned on board the \ac{pUAV}. This behavior is dangerous as it does not ensure the safety of the \ac{sUAV}, and methods to select rendezvous points, where can \ac{pUAV} correct this issues, are part of future work. Additionally, the large desired safety distances between the UAVs hindered the progress of the exploration, therefore, future work will focus on designing smaller \acp{UAV} with propeller protection.

The main purpose of the experiments was to prove that the proposed method is able to run in real time on the available hardware in real-world conditions with uncertainty. Even during the worst-case scenario, when there are no accessible \acp{POI} for \ac{pUAV}, the algorithms are fast enough to ensure smooth exploration.

\section{\textsc{Conclusion}}
We presented a novel method for unknown-space exploration using a heterogeneous team consisting of \acp{UAV} with different sizes and sensory equipment. 
The proposed approach utilizes a frontier-based method for generating Points of Interest, determining accessibility of the POIs using SphereMap, and solving the goal allocation problem using a minimum-cost flow technique. Collision-free paths to the selected POIs are generated and mutual distance between the UAVs is tracked to ensure safe flight.
The algorithms were tested in a complex simulation environment, and their performance was evaluated in a real-world experiment on board the UAVs, which had no external localization system or external computational resources available. These experiments not only showed the properties of the proposed method but also provided valuable insights into its real-world applicability and potential for further improvement.









\bibliographystyle{IEEEtran}
\bibliography{main.bib}

\end{document}